\documentclass{article}
\usepackage{times}
\usepackage{fullpage}
\usepackage[numbers]{natbib}
\usepackage{parskip}

\usepackage{mytable}
\usepackage{csvsimple}
\usepackage{xspace}
\usepackage{graphicx}
\usepackage{subfig}
\usepackage[utf8]{inputenc} 
\usepackage[T1]{fontenc}    
\usepackage{hyperref}       
\usepackage{url}            
\usepackage{booktabs}       
\usepackage{amsfonts}       
\usepackage{nicefrac}       
\usepackage{microtype}      
\usepackage[usenames,dvipsnames]{xcolor}
\renewcommand{\paragraph}[1]{\vspace{-0mm}\noindent{{\bf #1\ \ \ }}}
\usepackage{Definitions}
\usepackage{amsmath}
\usepackage{amssymb}
\usepackage{mathtools}
\usepackage{amsthm}
\usepackage{bm}
\usepackage[shortlabels,inline]{enumitem}
\setlist{nosep}
\usepackage{soul}
\usepackage{adjustbox}
\usepackage{multirow}
\usepackage{bigdelim}
\usepackage{rotating}
\usepackage{wrapfig,booktabs}
\usepackage{hyperref}

\newcommand{\subsubheader}{\paragraph}

\newcommand{\ad}{\textcolor{black}}
\usepackage[skip=.2ex plus.1ex minus.1ex]{caption}
\makeatletter
\g@addto@macro{\normalsize}{%
\setlength{\abovedisplayskip}{2pt plus1pt}%
\setlength{\abovedisplayshortskip}{2pt plus1pt}%
\setlength{\belowdisplayskip}{2pt plus1pt}%
\setlength{\belowdisplayshortskip}{2pt plus1pt}}
\let\c@table\c@figure
\makeatother
\newcommand{\our}{IsoNet++\xspace}
\newcommand{\ournode}{\our\ (Node)\xspace}
\newcommand{\ournodeshort}{IsoNet++\  (Node)\xspace}

\newcommand{\ouredge}{IsoNet++ (Edge)\xspace}
\newcommand{\ouredgeshort}{IsoNet++ (Edge)\xspace}

\newcommand\scc[1]{{\color{black} #1}}
\definecolor{cobalt}{rgb}{0.0, 0.28, 0.67}

\usepackage[capitalize,noabbrev]{cleveref}

\usepackage[rightComments=false,commentColor=blue,
beginComment=\quad//~,beginLComment=/*~, endLComment=~*/]{algpseudocodex}

\def\ztitle{Iteratively Refined Early Interaction Alignment for Subgraph Matching based Graph Retrieval}
\title{\ztitle}
\author{%
Ashwin Ramachandran$^{1*}$ \quad Vaibhav Raj$^{2*}$ \quad Indrayumna Roy$^2$ \\  {Soumen Chakrabarti}$^2$ \quad  {Abir De}$^2$ \\
$^1$UC San Diego \quad $^2$IIT Bombay\\
\texttt{ashwinramg@ucsd.edu}\\
\texttt{\{vaibhavraj, indraroy15, soumen, abir\}@cse.iitb.ac.in}\\
}

\date{}

\begin{document}

\maketitle
\def\thefootnote{*}\footnotetext{Equal contribution. Ashwin Ramachandran did this work while at IIT Bombay.}\def\thefootnote{\arabic{footnote}}

\begin{abstract}
Graph retrieval based on subgraph isomorphism has several real-world applications such as scene graph retrieval, molecular fingerprint detection and circuit design. \ad{\citet{RoyVCD2022IsoNet} proposed  IsoNet, a late interaction model for subgraph matching, which first computes the node and edge embeddings of each graph independently of paired graph  and then computes a trainable alignment map.}
Here, we present \our, an early interaction graph neural network (GNN), based on several technical innovations.
First, we compute embeddings of all nodes by passing messages within and across the two input graphs, guided by an \emph{injective alignment} between their nodes.
Second, we update this alignment in a lazy fashion over multiple \emph{rounds}. 
Within each round, we run a layerwise GNN from scratch, based on the current state of the alignment.
After the completion of one round of GNN, we use the last-layer embeddings to update the alignments, and proceed to the next round.
Third, \our{} incorporates a novel notion of node-pair partner interaction.
Traditional early interaction computes attention between a node and its potential partners in the other graph, the attention then controlling messages passed across graphs.
In contrast, we consider \emph{node pairs} (not single nodes) as potential partners.
Existence of an edge between the nodes in one graph and non-existence in the other provide vital signals for refining the alignment.
Our experiments on several datasets
show that the alignments get progressively refined with successive rounds,
resulting in significantly better retrieval performance than existing methods.
We demonstrate that all three innovations contribute to the enhanced accuracy. Our code and datasets are publicly available at \href{https://github.com/structlearning/isonetpp}{\textcolor{blue}{https://github.com/structlearning/isonetpp}}. 
\end{abstract}

\section{Introduction}
\label{sec:Intro}
In graph retrieval based on subgraph isomorphism, the goal is to identify a subset of graphs from a corpus,
denoted $\set{G_c}$, wherein each 
retrieved graph contains a subgraph isomorphic to a given query graph~$G_q$.  
Numerous real-life applications, \eg, molecular fingerprint detection~\cite{cereto2015molecular}, scene graph retrieval~\cite{johnson2015image}, circuit design~\cite{ohlrich1993subgemini} and frequent subgraph mining~\cite{yan2004graph}, can be formulated using subgraph isomorphism.
Akin to other retrieval systems,   the key challenge is to efficiently score corpus graphs against queries.

Recent work on neural graph retrieval~\cite{simgnn, graphsim, gotsim, gmn, neuromatch, RoyVCD2022IsoNet, egsc, eric} has shown significant promise. Among them, \citet[\neuromatch]{neuromatch} and \citet[\isonet]{RoyVCD2022IsoNet} focus specifically on subgraph isomorphism. They employ graph neural networks (GNNs) to obtain embeddings of query and corpus graphs and compute the relevance score using a form of order embedding~\citep{VendrovKFU2015OrderEmbeddings}. In addition, \isonet\ also approximates an \emph{injective alignment} between the query and corpus graphs.  These two models operate in a \emph{late interaction} paradigm, where the representations of the query and corpus graphs are computed independent of each other. In contrast, GMN \citep{gmn} is a powerful \emph{early interaction} network for graph matching, where GNNs running on $G_q$ and $G_c$ interact with each other at every layer.

Conventional wisdom suggests that early interaction is more accurate (even if slower) than late interaction, but GMN was outperformed by \isonet. This is because of the following reasons. 
\begin{enumerate*}[(1)]
\item GMN does not explicitly infer any alignment between $G_q$ and~$G_c$.  The graphs are encoded by two GNNs that interact with each other at every layer, mediated by attentions from each node in one graph on nodes in the other. These attentions are functions of node embeddings, so they change from layer to layer. While these attentions may be interpreted as approximate alignments, they induce at best non-injective mappings between nodes.

\item   In principle, one wishes to propose a consistent alignment across all layers. However, GMN's attention based `alignment' is updated in every layer.

\item GMN uses a standard GNN that is known to be an over-smoother~\citep{Rusch2023GnnOverSmoothing, Wenkel2022OversmoothScattering}. Due to this, the attention weights (which depend on the over-smoothed node representations) also suffer from oversmoothing.
\end{enumerate*}
These limitations raise the possibility of a \emph{third} approach based on early interaction networks, enabled with explicit alignment structures, that have the potential to outperform both GMN and \isonet.

\subsection{Our contributions}
\label{subsec:contrib}

We present \our, an early interaction network for subgraph matching that maintains a chain of explicit, iteratively refined, injective, approximate alignments between the two graphs.

\paragraph{Early interaction GNNs with alignment refinement}
We design early interaction networks for scoring graph pairs, that ensure the node embeddings of one graph are influenced by both its paired graph and the alignment map between them.
In contrast to existing works, we model alignments as an explicit ``data structure''.
An alignment can be defined between either nodes or edges, thus leading to two variants of our model: \ournode and \ouredge.
Within \our, we maintain a sequence of such alignments and refine them using GNNs acting on the two graphs. These alignments mediate the interaction between the two GNNs. 
In our work, we realize the alignment as a doubly stochastic approximation to a permutation matrix, which is an injective mapping by design.

\paragraph{Eager or lazy alignment updates}
In our work, we view the updates to the alignment maps as a form of gradient-based updates in a specific quadratic assignment problem or asymmetric Gromov-Wasserstein (GW) distance minimization~\cite{peyre2016gromov,lawrence2019gromov}.
The general form of \our{} allows updates that proceed lockstep with GNN layers (\emph{eager} layer-wise updates), but it also allows \emph{lazy} updates.  Specifically, \our{} can 
perform $T$ \emph{rounds} of updates to the alignment, each round including $K$ \emph{layers} of GNN message passing. 
During each round, the alignment is held fixed across all propagation layers in GNN. At the end of each round, we update
the alignment  by feeding the node embeddings into a neural Gumbel-Sinkhorn soft permutation generator~\cite{cuturi,Mena+2018GumbelSinkhorn,sinkhorn1967concerning}.

\paragraph{Node-pair partner interaction between graphs}
The existing remedies to counter oversmoothing~\citep{Cohenkarlik2020RecurrentPermutationInvariance, RoyDC2021permgnn, Wenkel2022OversmoothScattering} entail extra computation; but they may be expensive in an early-interaction setting. Existing early interaction models like~\cite{gmn} perform  \textit{node partner interaction}; interactions are constrained to occur between a node and it's \textit{partner}, the node in the paired graph aligned with it. Instead, we perform \textit{node-pair partner} interaction; the interaction is expanded to include the \textit{node-pairs} (or edges) in the paired graph that correspond to node-pairs containing the node. Consequently, the embedding of a node not only depends on nodes in the paired graph that align with it, but also captures signals from nodes in the paired graph that are aligned with its neighbors.

\paragraph{Experiments} The design components of \our\ and their implications are subtle ---
we report on extensive  experiments that tease out their effects.
Our experiments on real world datasets show that, \our\ outperforms several state-of-the-art methods for graph retrieval by a substantial margin.
Moreover, our results suggest 
that capturing information directly from node-pair partners can improve representation learning, as compared to taking information only from node partner.

\section{Preliminaries}
\label{eq:Prelim}

\paragraph{Notation} Given 
graph $G=(V,E)$, we use $\nbr(u)$ to denote the neighbors of a node $u\in V$. We use $u\to v$ to indicate a message flow from node $u$ to node $v$. Given a set of corpus graphs $C=\set{G_c}$ and a query graph  $G_q$, we denote $y(G_c \given G_q)$ as the  binary  relevance label of $G_c$ for $G_q$. 
Motivated by several real life applications like 
substructure search in molecular graphs~\cite{ehrlich2012systematic}, object search in scene graphs~\cite{johnson2015image}, and text entailment~\cite{lai2017entail}, 
we consider subgraph isomorphism to significantly influence the relevance label, similar to previous works~\cite{neuromatch,RoyVCD2022IsoNet}.
Specifically, $y(G_c \given G_q) =1$ when  $G_q$ is a subgraph of $G_c$, and $0$ otherwise.
We define $C_{q+} \subseteq C$ as the set of corpus graphs that are relevant to $G_q$ and set $C_{q-}=C\cp C_{q+}$. Mildly overloading notation, we use $\Pb$ to indicate a `hard' (0/1) permutation matrix or its `soft' doubly-stochastic relaxation.   $\Bcal_n$ denotes the set of  all $n\times n$ doubly stochastic matrices, and $\Pi_{n}$ denotes the set of all $n\times n$ permutation matrices. 

\paragraph{IsoNet~\cite{RoyVCD2022IsoNet}}
Given a graph $G=(V,E)$, IsoNet uses a GNN, which initializes node representations $\{ \hb_0(u): u\in V \}$
using node-local features.  Then, messages are passed between neighboring nodes in $K$ \emph{propagation  layers}. In the $k$th layer, a node $u$ receives messages from its neighbors, aggregates them, and then combines the result with its state after the $(k-1)$th layer:
\begin{align}
\textstyle \hb_k (u) = \comb_{\theta}\big(\hb_{k-1} (u),  
\sum_{v\in \nbr(u)}\set{\msg_{\theta}(\hb_{k-1} (u), \hb_{k-1} (v))} \big).\label{eq:naivegnn}
\end{align}
Here, $\msg_{\theta} (\cdot) $ and $ \comb_{\theta}(\cdot, \cdot)$ are suitable networks with parameters collectively called~$\theta$. 
Edges may also be featurized and influence the messages that are aggregated~\citep{marcheggiani-titov-2017-encoding}.
The node representations at the final propagation layer $K$ can be collected into the matrix $\Hb = \{ \hb_K(u)\given u\in V\}$. 
Given a node $u\in G_q$ and a node $u'\in G_c$, we denote the embeddings of $u$ and $u'$ after
the propagation layer $k$ as $\hq _k (u)$ and $\hc _k(u')$ respectively.
$\Hq$ and $\Hc$ denote the $K$th-layer node embeddings of $G_q$ and $G_c$, collected into matrices. Note that, here 
the set of vectors $\Hq$ and $\Hc$ do not dependent on $G_c$ and $G_q$. In the end, IsoNet compares these embeddings to compute the distance $    \Delta (G_c\given G_q)$, which is  inversely related to $\hat{y}(G_c \given G_q)$.
\begin{align}
    \Delta (G_c\given G_q)  =  \textstyle\sum_{u,i}\text{ReLU}[\Hc-\Pb \Hq][u,i] \label{eq:isonode}
\end{align}
Since subgraph isomorphism entails an asymmetric relevance, we have: $\Delta ( G_c \given G_q) \neq\Delta ( G_q \given G_c)$.
IsoNet also proposed another design of $\Delta$, where it replaces the node embeddings with edge embeddings and node alignment matrix with edge alignment matrix in Eq.~\eqref{eq:isonode}.

In an \textbf{early} interaction network, $\Hq$ depends on $G_c$ and $\Hc$ depends on $G_q$ for any given $(G_q, G_c)$ pair. Formally, one should write $\Hb^{(q\given c)}$ and $\Hb^{(c\given q)}$ instead of $\Hq$ and $\Hc$ respectively for an early interaction network, but for simplicity, we will continue using $\Hq$ and $\Hc$.

\paragraph{Our goal} Given a set of corpus graphs $C=\set{G_c \given c \in [|C|]}$, our high-level goal is to build a graph retrieval model so that, given a query $G_q$, it can return the corpus graphs $\set{G_c}$ which are relevant to $G_q$. To that end, we seek to develop
\begin{enumerate*}[(1)]
\item a GNN-based early interaction model, and
\item an appropriate distance measure $\Delta(\cdot \given \cdot)$, so that $\Delta(\Hc \given \Hq)$ is an accurate predictor of $y(G_c\given G_q)$, at least to the extent that $\Delta(\cdot|\cdot)$ is effective for ranking candidate corpus graphs in response to a query graph.
\end{enumerate*}

\section{Proposed early-interaction GNN with multi-round alignment refinement}
\label{sec:Design}

In this section, we first write down the subgraph isomorphism task as an instance of the quadratic assignment problem (QAP) or the Gromov-Wasserstein (GW) distance optimization task. Then, we design \our, by building upon this formulation.

\subsection{Subgraph isomorphism as Gromov-Wasserstein distance optimization}

\paragraph{QAP or GW formulation with asymmetric cost}
We are given a graph pair $G_q$ and $G_c$ padded with appropriate number of nodes to ensure $|V_q|=|V_c|=n$ (say). Let their adjacency matrices be $\Ab_q, \Ab_c \in \set{0,1}^ {n \times n}$. Consider the family of hard permutation matrices $\Pb\in \Pi_{n}$ where $\Pb[u,u']=1$ indicates $u\in V_q$ is ``matched'' to $u'\in V_c$. 
Then, $G_q$ is a subgraph  of $G_c$, if for some permutation matrix $\Pb$, the matrix  $\Ab_q$ is covered by $\Pb\Ab_c \Pb^{\top}$, \ie, for each pair $(u,v)$, whenever we have $\Ab_{q}[u,v]=1$, we will also have $\Pb\Ab_c \Pb^{\top}[u,v]=1$.  This condition can be written as $\Ab_q \le \Pb \Ab_c \Pb^{\top}$.
We can regard a deficit in coverage as a cost or distance:
\begin{align}
\operatorname{cost}(\Pb; \Ab_q, \Ab_c) & =  \textstyle
\sum_{u\in [n],v\in[n]} \left[\big(\Ab_q-\Pb \Ab_c \Pb^{\top}\big)_+\right] [u,v] \label{eq:hard-score}\\
& = \textstyle 
    \sum_{u,v\in [n]}\sum_{u',v'\in [n]}(\Ab_q[u,v]-\Ab_c[u',v'])_+ \; \Pb[u,u']\; \Pb[v,v'] \label{eq:qap}
\end{align}
Here, $[\cdot]_+= \max\set{\cdot,0}$ is the ReLU function, applied elementwise. The function $\mathrm{cost}(\Pb; \Ab_q, \Ab_c)$ can be driven down to zero using a suitable choice of $\Pb$ iff $G_q$ is a subgraph of $G_c$. This naturally suggests the relevance distance
\begin{align}
    \Delta (G_c \given G_q) & =  \min_{\Pb \in \Pi_{n}}   \mathrm{cost}(\Pb; \Ab_q, \Ab_c) \label{eq:costOpt}
\end{align}
\citet{lawrence2019gromov} demonstrate that this QAP is a realization of the Gromov-Wassterstein distance minimization in a graph setting.

\paragraph{Updating $\Pb$ with projected gradient descent}  
As shown in~\citet{benamou2015gromov,peyre2016gromov,lawrence2019gromov}, one approach is to first relax $\Pb$ into a doubly stochastic matrix, which serves as a continuous approximation of the discrete permutation, and then update it using projected gradient descent (PGD). Here, the soft permutation $\Pb_{t-1}$ is updated to $\Pb_t$ at time-step $t$ by solving the following linear optimal transport (OT) problem, regularized with the entropy of $\set{\Pb[u,v]\given u,v\in [n]}$ with a temperature $\tau$. 
\begin{align}
    \Pb_{t} \leftarrow \argmin _{\Pb \in \Bcal_{n}} \textrm{Trace}\left(\Pb^{\top}\nabla_{\Pb} \textrm{cost}(\Pb;\Ab_q,\Ab_c) \big|_{\Pb = \Pb_{t-1}}\right) + \tau \sum_{u,v}\Pb[u,v] \cdot \log \Pb[u,v] . \label{eq:PGWupdate}
\end{align}
Such an OT problem is solved  using  the iterative Sinkhorn-Knopp algorithm~\cite{cuturi,sinkhorn1967concerning,Mena+2018GumbelSinkhorn}. Similar to other combinatorial optimization problems on graphs, a QAP~\eqref{eq:qap} does not capture the coverage cost in the presence of dense node or edge features,
where two nodes or edges may exhibit graded degrees of similarity represented by continuous values. Furthermore, the binary values of the adjacency matrices result in inadequate gradient signals in $\nabla_{\Pb} \mathrm{cost}(\cdot)$.   
Additionally, the computational bottleneck of solving a fresh OT problem in each PGD step introduces a significant overhead, especially given the large number of pairwise evaluations required in typical learning-to-rank setups.

\begin{figure}
    \centering
 \includegraphics[width=0.96\textwidth]{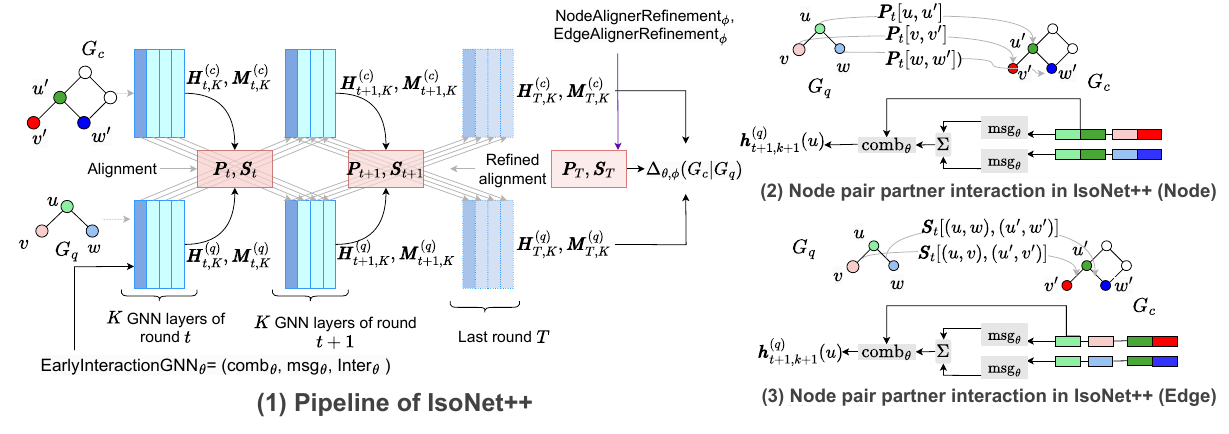}
\caption{%
Overview of \our. Panel~(a) shows the pipeline of \our. Given a graph pair $(G_q,G_c)$, we execute $T$ \emph{rounds}, each consisting of $K$ GNN \emph{layer} propagations. 
After a round $t$, we use the node embeddings to update the node alignment $\Pb=\Pb_{t}$ from its previous estimate $\Pb=\Pb_{t-1}$.
Within each round $t\in[T]$, we compute the node embeddings of $G_q$ by gathering signals from $G_c$ and vice-versa, using GNN embeddings in the previous round and the
node-alignment map $\Pb_{t}$. The alignment $\Pb_{t}$ remains consistent 
across all propagation layers $k\in[K]$ and is updated at the end of round $t$. 
Panel~(b) shows our proposed node pair partner interaction in \ournode. When computing the message value of the node pair $(u,v)$, we also feed the node embeddings of the partners $u'$ and $v'$ in addition to the embeddings of the pairs $(u,v)$, where  $u'$ and $v'$ is  approximately aligned with $u$ and $v$, respectively \ad{(when converted to soft alignment, $u',v'$ need not be neighbors)}. 
Panel~(c) shows the node pair partner interaction in \ouredge. In contrast to \ournode, here we feed the information from the message value of the partner pair $(u',v')$ instead of their node embeddings into the message passing network $\msg_{\theta}$.
}  
\label{fig:demo}
\end{figure}
\subsection{Design of \ournode}
\label{subsec:our_node_design}

Building upon the insights from the above GW minimization~\eqref{eq:hard-score} and the successive refinement step~\eqref{eq:PGWupdate},  we build \ournode, the first variant of our proposed early interaction model.

\begin{figure}[ht]
\centering
\captionsetup[subfloat]{justification=centering} 

\begin{subfloat}[{No Interaction\\(IsoNet)}] {
\centering

\scalebox{0.30}{
 \includegraphics[width=0.75\textwidth]{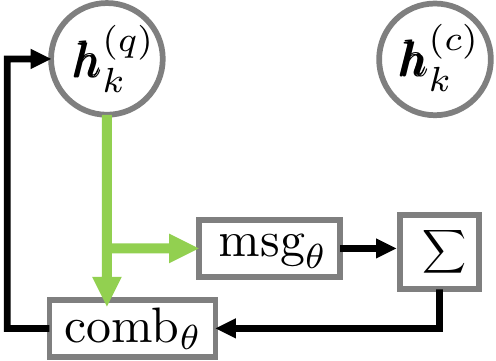}
 }}
\end{subfloat}
\hspace{15pt}      
\centering
\begin{subfloat}[{Node Pair Interaction\\ (GMN)}] {
\centering
\scalebox{0.30}{
 \includegraphics[width=0.75\textwidth]{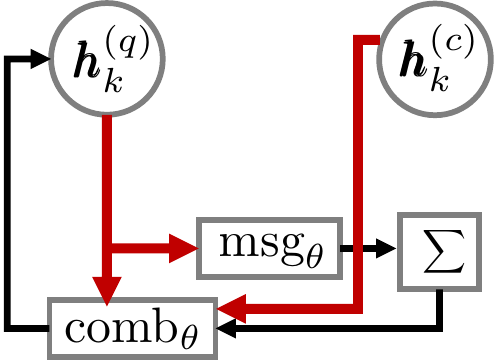}
}}
\end{subfloat}
\hspace{15pt}
\centering
\begin{subfloat}[{Node Pair Partner\\ Interaction  (\our)}] {
\centering
\scalebox{0.30}{
 \includegraphics[width=0.75\textwidth]{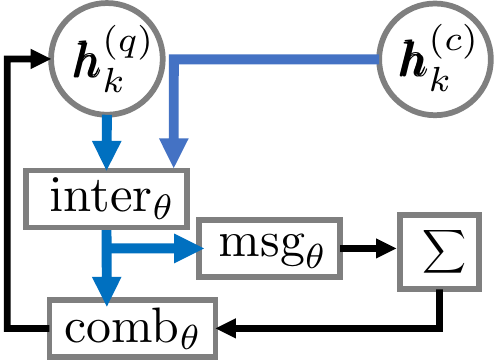}
}}
\end{subfloat}
\caption{Illustration of the three interaction modes. IsoNet has no/late interaction between $\hq$ and $\hc$. \our and \gmn\ allow interaction between the representations of the query and corpus nodes. Under \textbf{node pair interaction}, the individual node embeddings $\hq$ are used for message passing directly, thereby exposing them only to their neighbors. In the corresponding $\textstyle\comb_{\theta}$ step, nodes interact only with their respective partners, therefore missing out on information from the partners of its neighbors. However, under \textbf{node pair partner interaction}, the representation of a node is combined with that of its partner(s) first, using the $\inter_\theta$ block to obtain $\zq$~\eqref{eq:inter-1}, which is used for message passing. Thus, when interacting with its neighbors, a node also gets information from the partners of its neighbors.}
\label{fig:interaction}

\end{figure}

\subsubheader{Node-pair partner interactions between graphs} 
For simpler exposition, we begin by describing  a synthetic scenario, where $\Pb$ is a hard node permutation matrix, which induces the alignment map as a bijection $\pi:V_q\to V_c$, so that $\perm{a}=b$ if $\Pb[a,b]=1$. We first initialize layer $k=0$ embeddings as $\hq_0 (u) =  \init_{\theta}(\feat(u) )$ using a neural network $\init_{\theta}$. (Throughout, $\hc_k(u)$ are treated likewise.)
Under the given alignment map $\pi$, a simple early interaction model would update the node embeddings as follows:
\begin{align}
\hq_{k+1}(u) &= \textstyle\comb_{\theta}\left(
\hq_{k}(u), \; \sum_{v\in\nbr(u)} \msg_{\theta}(\hq_{k}(u), \hq_{k}(v)), \;
 \hc_{k}(\perm{u}) \right) \label{eq:gmnupdate}
\end{align}
In the above expression, the update layer uses representation of the partner node $u'\in V_c$ during the message passing step, to compute $\hb^{(q)}_{k+1}(u)$,  the embedding of node $u \in V_q$.~\citet{gmn} use a similar update protocol, by approximating 
$\hb_k ^{(c)} (\perm{u}) = \sum_{u'\in V_c}a^{(k)}_{u'\to u} \hb_k ^{(c)}(u')$, where  $a^{(k)}_{u'\to u}$ is the $k$th layer attention from $u\in V_q$ to potential partner $u'\in V_c$, with $\sum_{u'\in V_c} a^{(k)}_{u'\to u}=1$.
Instead of regarding only nodes as potential partners, \our{} will regard \emph{node pairs} as partners.  Given $(u,v)\in E_q$, the partners $(\perm{u}, \perm{v})\in E_c$ should then greatly influence the intensity of assimilation of $\hc_k(u')$ into $\hc_{k+1}(u)$.  The first  key innovation in \our{} is to replace \eqref{eq:gmnupdate} to recognize and implement this insight:
\begin{multline}
\hq_{k+1}(u) = 
\comb_\theta\Big(
\matt{\hq_k(u), \hc_k(\perm{u})}, \\[-1ex]
\textstyle \sum_{v\in\nbr(u)} 
\msg_\theta\big( \matt{\hq_k(u), \hc_k(\perm{u})},
\matt{\hq_k(v), \hc_k(\perm{v})} \big) \Big)  \label{eq:einupdate}
\end{multline}
Embeddings $\hc_{k+1}(u')$ for nodes $u'\in V_c$ are updated likewise in a symmetric manner.  The network $\msg_\theta$ is provided embeddings from partners $\perm{u}, \perm{v}$ of $u,v\in V_q$ --- this allows $\hb^{(\bullet)}_{k+1}(u)$ to capture information from all nodes in the paired graph, that match with the $(k+1)$-hop neighbors of~$u$. We schematically illustrate the interaction between the paired graphs in \isonet, \gmn and \our in Figure~\ref{fig:interaction}.

\subsubheader{Multi-round lazy refinement of node alignment}
In reality, we are not given any alignment map $\pi$. This motivates   our second key innovation   beyond prior models~\cite{simgnn, gmn, neuromatch, RoyVCD2022IsoNet}, where we decouple GNN layer propagation from updates to~$\Pb$.  To achieve this, \ournode{} executes $T$ \emph{rounds}, each consisting of $K$ \emph{layer} propagations in both GNNs.
At the end of each round $t$, we refine the earlier alignment $\Pb_{t-1}$ to the next estimate~$\Pb_t$, which will be used in the next round.
Henceforth, we will use the double subscript $t,k$ instead of the single subscript $k$ as in traditional GNNs.
We denote the node embeddings at layer $k$ and round $t$ by $\hb^{(q)}_{t,k}(u),  \hb^{(c)}_{t,k}(u') \in \RR^{\dim_h}$ for $u\in V_q$ and $u'\in V_c$, which are (re-)initialized with node features $\hb^{\bullet} _{t,0}$ for each round~$t$. We gather these into matrices
\begin{align}
\Hqtk{t}{k} = [ \hb^{(q)}_{t,k}(u)\given u\in V_q ] \in \RR^{n\times \text{dim}_h}
\quad \text{and} \quad
\Hctk{t}{k}  = [ \hb^{(c)}_{t,k} (u')\given u'\in V_c ] \in \RR^{n\times \text{dim}_h}.
\label{eq:hqtk}
\end{align}
$\Pb$ no longer remains an oracular hard permutation matrix, but becomes a doubly stochastic matrix indexed by rounds, written as~$\Pb_t$.
At the end of round $t$, a differentiable \emph{aligner} module takes $\Hqtk{t}{K}$ and $\Hctk{t}{K}$ as inputs and outputs a doubly stochastic node alignment (relaxed permutation) matrix $\Pb_{t}$ as follows:
\begin{align}
\Pb_{t}&  = \operatorname{NodeAlignerRefinement}_{\phi}\big(\Hqtk{t}{K},\Hctk{t}{K} \big) \\
& \begin{aligned}
  &  = \operatorname{GumbelSinkhorn}\left( \operatorname{LRL}_\phi(\Hqtk{t}{K})
  \operatorname{LRL}_\phi(\Hctk{t}{K}) ^\top \right) \in \Bcal_n \label{eq:Pt+1}
\end{aligned}
\end{align}
In the above expression, $\mathrm{GumbelSinkhorn}(\bullet)$ performs iterative Sinkhorn normalization on the input matrix added with Gumbel noise~\cite{Mena+2018GumbelSinkhorn}; LRL$_\phi$ is a neural module consisting of two linear layers with a ReLU activation after the first layer. 
As we shall see next, $\Pb_{t}$ is used to gate messages flowing \emph{across} from one graph to the other during round $t+1$, i.e., while computing $\Hqtk{t+1}{1{:}K}$ and $\Hctk{t+1}{1{:}K}$.  The soft alignment $\Pb_{t}$ is kept frozen for the duration of all layers in round~$t+1$.  $\Pb_{t}[u,u']$ may be interpreted as the probability that $u$ is assigned to $u'$, which naturally requires that $\Pb_{t}$ should be row-equivariant (column equivariant) to the shuffling of the node indices of $G_q$ ($G_c$). As shown in Appendix~\ref{app:model}, the above design choice~\eqref{eq:Pt+1} ensures this property.

\subsubheader{Updating node representation using early-interaction GNN}
Here, we describe the early interaction GNN for the query graph $G_q$. 
The GNN on the corpus graph $G_c$ follows the exact same design and is deferred to Appendix~\ref{app:multi-round-for-corpus}.
In the initial round ($t=1$), since there is no prior alignment estimate $\Pb_{t=0}$, we employ the traditional late interaction GNN~\eqref{eq:naivegnn} to compute all layers $\Hqtk{1}{1:K}$ and $\Hctk{1}{1:K}$ separately.  These embeddings are then used to estimate $\Pb_{t=1}$ using Eq.~\eqref{eq:Pt+1}.  For subsequent rounds ($t>1$), given embeddings $\Hqtk{t}{1:K}$, and the alignment estimate matrix $\Pb_{t}$, we run an early interaction GNN from scratch. We start with a fresh initialization of the node embeddings as before; i.e.,
$\hq_{t+1,0} (u) =  \init_{\theta}(\feat(u))$.
For each subsequent propagation layer $k+1$ ($k\in[0,K-1]$),
we approximate \eqref{eq:einupdate} as follows.
We read previous-round, same-layer embeddings $\hc_{t,k}(u')$ of nodes $u'$ from the other graph $G_c$, incorporate the alignment strength $\Pb_t[u,u']$, and aggregate these to get an intermediate representation of $u$ that is sensitive to $\Pb_t$ and~$G_c$.
\begin{align}
    &\zq_{t+1,k}(u) = \textstyle \inter_{\theta}\Big(\hq_{t+1,k}(u), \sum_{u'\in V_c} \hc_{t,k}(u') \Pb_{t}[u,u']\Big)  \label{eq:inter-1}
\end{align}
Here, $\inter_{\theta}$ is a neural network that computes interaction between the graph pairs; $\zq_{t+1,k}(u)$
provides a soft alignment guided representation
of $[\hq_{k}(u), \hc_{k}(\perm{u})]$ in Eq.~\eqref{eq:einupdate}, which can be relaxed as:
\begin{align}
\hqtk{t+1}{k+1}(u) &=  \textstyle\comb_{\theta}\Big(\zq_{t+1, k} (u),  
\sum_{v\in \nbr(u)} \msg_{\theta}(\zq_{t+1, k} (u), \zq_{t+1, k} (v)) \Big)     \label{eq:node-emb-n1}
\end{align}
In the above expression, we explicitly feed  $\zb^{(q)} _{t+1,k}(v), v\in\nbr(u)$ in the $\msg_{\theta}$ network, capturing embeddings of nodes in the corpus $G_c$ aligned with the \emph{neighbors} of node $u\in V_q$ in $\hqtk{t+1}{k+1} (u)$. This allows the model to perform node-pair partner interaction. Instead, if we were to feed only $\hb^{(q)} _{t+1,k}(u)$ into the $\msg_{\theta}$ network, then it would only  perform node partner interaction. In this case, the computed embedding for $u$ would be based solely on signals from nodes in the paired graph that directly correspond to $u$, therefore missing additional context from other neighbourhood nodes.

\subsubheader{Distant supervision of alignment}
Finally, at the end of $T$ rounds, we express the relevance distance $\Delta(G_c \given G_q)$ as a soft distance between the set $\Hqtk{T}{K}=[\hqtk{T}{K}(u) \given u\in V_q]$ and $\Hctk{T}{K}=[\hctk{T}{K}(u') \given u'\in V_c]$, measured as
\begin{align}
\Delta_{\theta,\phi}(G_c \given G_q) &= 
\textstyle\sum_u \sum_d \operatorname{ReLU}(
\Hqtk{T}{K}[u, d] - (\Pb_{T} \Hctk{T}{K})[u, d] )
\label{eq:node-score}
\end{align}
Our focus is on graph retrieval applications.  It is unrealistic to assume direct supervision from a gold alignment map~$\Pb^*$.  Instead, training query instances are associated with pairwise preferences between two corpus graphs, in the form $\langle G_q, G_{c+}, G_{c-}\rangle$, meaning that, ideally, we want $\Delta_{\theta,\phi}(G_{c-}|G_q) \ge \gamma + \Delta_{\theta,\phi}(G_{c+}|G_q)$, where $\gamma>0$ is a margin hyperparameter.  This suggests a minimization of the standard hinge loss as follows:
\begin{align} 
\textstyle\min_{\theta,\phi} \sum_{q\in Q}\sum_{c+\in C_{q+},c-\in C_{q-}}[\gamma+\Delta_{\theta,\phi}(G_{c+}\given G_q) - \Delta_{\theta,\phi}(G_{c-}\given G_q)]_+
\label{eq:rankingloss}
\end{align}
This loss is back-propagated to train model weights $\theta$ in $\comb_\theta, \inter_\theta, \msg_\theta$ and weights $\phi$ in the Gumbel-Sinkhorn network.

\paragraph{Multi-layer eager alignment variant}
Having set up the general multi-round framework of \our, we introduce a structurally simpler variant that updates $\Pb$ eagerly after every layer, eliminating the need to  
re-initialize node embeddings every time we update~$\Pb$.  The eager variant retains the benefits of node-pair partner interactions, while ablating \our{} toward GMN.  Updating $\Pb$ via Sinkhorn iterations is expensive compared to a single GNN layer. \scc{In practice, we see a non-trivial tradeoff between computation cost, end task accuracy, and the quality of our injective alignments, depending on the value of $K$ for eager updates, and the values $(T,K)$ for lazy updates.} 
 Formally,   $\Pb_{k}$ is updated across layers as follows:
\begin{align}
     \Pb_{k} & =\operatorname{NodeAlignerRefinement}_{\phi}\big(\Hq_{k},\Hc_{k} \big) \\
  & = \operatorname{GumbelSinkhorn}\left( \operatorname{LRL}_\phi(\Hq_{k})
  \operatorname{LRL}_\phi(\Hc_{k}) ^\top \right). \label{eq:Pt+1-multilayer}
\end{align}
 We update the GNN embeddings, layerwise,  as follows:
\begin{align}
&    \zq_{ k}(u) = \textstyle \inter_{\theta}\Big(\hq_{k}(u), \sum_{u'\in V_c} \hc_{k}(u') \Pb_{k}[u,u']\Big), \label{eq:inter-1-multi-layer}\\   
 &   \hb^{(q)}_{ k+1 }(u) = \textstyle\comb_{\theta}\Big(\zq_{ k} (u),  
\sum_{v\in \nbr(u)} \msg_{\theta}(\zq_{k} (u), \zq_{k} (v)) 
\label{eq:inter-2-multi-layer}\Big)
\end{align}
\paragraph{Analysis of computational complexity}
Here, will compare the performance of \isonode~\cite{RoyVCD2022IsoNet} with \ournodelayer and \ournoderound for graphs with $|V|$ nodes. For \ournodelayer and \isonode, we assume $K$ propagation steps and for \ournoderound, $T$ rounds, each with $K$ propagation steps.

\emph{---\isonode:} The total complexity is $O(|V|^2 + K|E|)$, computed as follows:
\begin{enumerate*}[leftmargin=5mm, label=\textbf{(\arabic*)}]
    \item Initialization of layer embeddings at layer $k=0$ takes $O(|V|)$ time.
    \item The node representation computation  incurs a complexity of $O(|E|)$ for each message passing step since it aggregates node embeddings across all neighbors.
    \item The computation of $\Pb$ takes $O(|V|^2)$ time.
\end{enumerate*}

\emph{---Multi-layer eager \our (Node):} The total complexity is $O(K|V|^2+K|E|+K|V|^2)=O(K|V|^2)$, computed as follows:
\begin{enumerate*}[leftmargin=5mm, label=\textbf{(\arabic*)}]
    \item Initialization (layer $k=0$) takes $O(|V|)$ time.
    \item The computation of intermediate embeddings $\zb^{(\bullet)}$ (Eq.~\ref{eq:inter-1-multi-layer}) involves the evaluation of the expression $\sum _{u'\in V _c} \hb^{(\bullet)} _{k}(u') \Pb _{k}[u,u']$ and hence admits a complexity of $O(|V|)$ for each node per layer. The total complexity for $K$ steps and $|V|$ nodes is thus $O(K|V|^2)$.
    \item Next, for each node in every layer, we compute $\hb ^{(\bullet)}_{k+1}$ (Eq.~\ref{eq:inter-2-multi-layer}) which gathers messages $\zb^{(\bullet)}$ from all its neighbors, contributing a total complexity of $O(K|E|)$.
    \item Finally, we update $\Pb  _k$ for each layer which has a complexity of $O(K|V|^2)$.
\end{enumerate*}

\emph{---Multi-round \our (Node):} Here, the key difference from the multi-layer version above is that the doubly stochastic matrix $\Pb _t$ from round $t$ is used to compute $\zb$ and the $K$-step-GNN runs in each of the $T$ rounds. This multiplies the complexity of steps 2 and 3 with $T$, raising it to $O(KT|V|^2+KT|E|)$. Matrix $\Pb _t$ is updated a total of $T$ times, which changes the complexity of step 4 to $O(T|V|^2)$. Hence, the total complexity is $O(KT|V|^2+T|V|^2+KT|E|) = O(KT|V|^2)$.

Hence, the complexity of IsoNet is $O(|V|^2+K|E|)$, multi-layer \our is $O(K|V|^2)$ and multi-round \our is $O(KT|V|^2)$. This increased complexity of the latter comes with the benefit of a significant performance boost, as our experiments suggest.

\subsection{Extension of \ournode{} to \ouredge}
\label{subsec:our_edge_design}

We now extend \ournode\ to \ouredge\ which uses explicit edge alignment for interaction across GNN and relevance distance surrogate.

\paragraph{Multi-round refinement of edge alignment} 
In \ouredge, we maintain a soft edge permutation matrix $\Sb$ which is frozen at $\Sb=\Sb_{t-1}$ within each round $t \in [T]$
and gets refined after every round $t$ as $\Sb_{t-1} \to \Sb_{t}$. Similar to \ournode, within each round $t$, GNN runs from scratch: it propagates messages across layers $k\in [K]$ and $\Sb_{t-1}$ assists it to capture cross-graph signals.
Here, in addition to node embeddings $\htk{t}{k}$, we also use edge embeddings
$\mqtk{t}{k}(e),\ \mctk{t}{k}(e') \in \RR^{\dim_m}$ at each layer $k$ and each round $t$, which capture the information about the subgraph $k\le K$ hop away from the edges $e$ and~$e'$.
Similar to Eq.~\eqref{eq:hqtk}, we define
    $\Mqtk{t}{k} = [\mqtk{t}{k}(e)]_{e \in E_q}, $ and $\Mctk{t}{k} = [\mctk{t}{k}(e')]_{e'\in E_c}$.
$\Mtk{t}{0}$ are initialized using the features of the nodes connected by the edges, and possibly local edge features.
Given the embeddings $\Mqtk{t}{K}$ and $\Mctk{t}{K}$ computed at the end of round $t$, an edge aligner module ($\operatorname{EdgeAlignerRefinement}_{\phi}(\bullet)$) takes these embedding matrices as input and outputs a soft edge permutation matrix $\Sb_t$, similar to the update of $\Pb_t$ in Eq.~\eqref{eq:Pt+1}. 
\begin{align}
\Sb_{t}&  = \operatorname{EdgeAlignerRefinement}_{\phi}\big(\Mqtk{t}{K},\Mctk{t}{K} \big) \\
& \begin{aligned}
  &  = \operatorname{GumbelSinkhorn}( \operatorname{LRL}_\phi(\Mqtk{t}{K})   \,\operatorname{LRL}_\phi(\Mctk{t}{K}) ^\top ) \label{eq:St+1}
\end{aligned}
\end{align}
Here, $\Mtk{t}{K}$ are appropriately padded to ensure that they have the same number of rows.

\paragraph{Edge alignment-induced early interaction GNN} For $t=1$, we start with a late interaction model using vanilla GNN~\eqref{eq:naivegnn} and obtain $\Sb_{t=1}$ using Eq.~\eqref{eq:St+1}. Having computed the edge embeddings $\mtk{t}{1:K}(\bullet)$ and node embeddings $\htk{t}{1:K}(\bullet)$ upto round $t$, we compute $\Sb_t$ and use it to build a fresh early interaction GNN for round $t+1$. To this end, we adapt the GNN guided by $\Pb_t$ in Eqs.~\eqref{eq:inter-1}--~\eqref{eq:node-emb-n1}, 
to the GNN guided by $\Sb_t$. We overload the notations for neural modules and different embedding vectors from \ournode, whenever their roles are similar.

Starting with the same initialization as in \ournode, we perform the cross-graph interaction guided by the soft edge permutation matrix $\Sb_t$, similar  to Eq.~\eqref{eq:inter-1}. Specifically, we use the embeddings of edges $\set{e'=(u',v')}\in E_c$, computed at layer $k$ at round $t$, which share soft alignments with an edge $e=(u,v)\in E_q$, to compute $\zq_{t+1,k}(e)$ and $\zq_{t+1,k}(e')$ as follows:
 \begin{align}
    &\zq_{t+1,k}(e) = \textstyle \inter_{\theta}\Big(\text{{$\mq_{t+1,k}(e),$}} \sum_{e'\in E_c} \mc_{t,k}(e') \Sb_{t}[e,e']\Big) \label{eq:inter-e1}
\end{align} 
Finally, we update the node embeddings $\htk{t+1}{k+1}$ for propagation layer $k+1$ as
\begin{align}
&\hspace{-4mm}  \hqtk{t+1}{k+1}(u)   =  \textstyle\comb_{\theta}\Big(\hqtk{t+1}{k} (u),  
\sum_{a\in \nbr(u)} \msg_{\theta}(\hqtk{t+1}{k} (u), \hqtk{t+1}{k} (a), \zq_{t+1,k}((u,a))) \Big)  \label{eq:node-emb-e1}
\end{align}
In this case, we perform the cross-graph interaction at the edge level rather than the node level. Hence,  $\msg_{\theta}$ acquires cross-edge signals separately as $\zb^{(\bullet)} _{t+1,k}$. Finally, we use $\htk{t+1}{k+1}$ and $\zb^{(\bullet)} _{t+1,k+1}$ to update $\mtk{t+1}{k+1}$ as follows:
\begin{align}
    \mqtk{t+1}{k+1}\big((u,v)\big) =\msg_{\theta}\Big(\hqtk{t+1}{k+1} (u), \hqtk{t+1}{k+1} (v), \zq_{t+1,k}((u,v))\Big)
\end{align}
Likewise, we develop $\mctk{t+1}{k+1}$ for corpus graph $G_c$. Note that $\mqtk{t+1}{k+1}
((u,v))$ captures signals not only from the matched pair $(u',v')$, but also 
signals from the nodes in $G_c$ which share correspondences with the neighbor nodes 
of $u$ and $v$. Finally, we pad zero vectors to $[\mqtk{T}{K}(e)]_{e \in E_q}$ and 
$[\mctk{T}{K}(e')]_{e'\in E_c}$ to build the matrices $\Mqtk{T}{K}$ and $\Mctk{T}{K}$ with same number of rows, 
which are finally used to compute the relevance distance 
\begin{align}
\Delta_{\theta,\phi}(G_c \given G_q) &= 
\textstyle\sum_u \sum_d \operatorname{ReLU}(
\Mqtk{T}{K}[e, d] - (\Sb_{T} \Mctk{T}{K})[e, d] ).
\label{eq:edge-score}
\end{align}
 
 \section{Experiments}
\label{sec:Expt}

We report on a comprehensive evaluation of \our{} on six real datasets and analyze the efficacy of the key novel design choices. In Appendix~\ref{app:expts}, we provide results of additional experiments.

\subsection{Experimental setup}

\paragraph{Datasets}
We use six real world datasets  in our experiments, \viz, \aids, \mutag, PTC-FM (\fm), PTC-FR (\fr), PTC-MM (\mm) and PTC-MR (\mr), which were also used in \cite{Morris+2020,RoyVCD2022IsoNet}. Appendix~\ref{app:details} provides the details about dataset generation and their statistics.

\paragraph{State-of-the-art baselines}
We compare our method against eleven state-of-the-art methods, \viz,
\begin{enumerate*}[label=(\arabic*)]
\item \graphsim~\cite{graphsim}
\item \gotsim~\cite{gotsim}, 
\item \simgnn~\cite{simgnn}, 
\item \egsc~\cite{egsc},
\item \hmn~\cite{zhang2021h2mn},
\item \neuromatch~\cite{neuromatch},
\item  \greed~\cite{greed},
\item \gen~\cite{gmn}, 
\item \gmn~\cite{gmn} 
\item \isonode~\cite{RoyVCD2022IsoNet},
and \item \isoedge~\cite{RoyVCD2022IsoNet}.   
\end{enumerate*}
Among them, \neuromatch, \greed, \isonode\ and \isoedge\ 
apply asymmetric hinge distances between query and corpus embeddings for $\Delta(G_c\given G_q)$, specifically catered towards subgraph matching, similar to our method in Eqs.~\eqref{eq:node-score} and~\eqref{eq:edge-score}.
\gmn\ and \gen\ use symmetric Euclidean distance between their (whole-) graph embeddings $\bm{g}^{(q)}$ (for query) and $\bm{g}^{(c)}$ (for corpus) as
$||\bm{g}^{(q)}-\bm{g}^{(c)}||$ in their paper~\cite{gmn}, which is not suitable for subgraph matching and therefore, results in poor performance.
Hence, we change it to $\Delta(G_c\given G_q)=
[\bm{g}^{(q)}-\bm{g}^{(c)}]_+$. The other methods first compute the graph embeddings, then fuse them using a neural network and finally apply a nonlinear function on the fused embeddings to obtain the relevance score. 

\paragraph{Training and evaluation protocol}
Given a fixed corpus set $C$, we split the query set $Q$ into $60\%$ training, $15\%$ validation and $25\%$ test set. We train all the models on the training set by minimizing a ranking loss~\eqref{eq:rankingloss}.
During the training of each model, we use five random seeds.
Given a test query $q'$,  we rank the corpus graphs $C$ in the decreasing order of $\Delta_{\theta,\phi}(G_c\given G_{q'})$ computed using the trained model. We evaluate the quality of the ranking by measuring \AP{} (AP) and HITS@20, described in Appendix~\ref{app:details}. Finally, we report mean average precision (MAP) and mean HITS@20, across all the test queries. 
By default, we set the number of rounds $T=3$, the number of propagation layers in GNN $K=5$. In Appendix~\ref{app:details}, we discuss the baselines, hyperparameter setup and the evaluation metrics in more detail.

\subsection{Results}
\begin{table}[t]
\centering\adjustbox{max width=.95\hsize}{\tabcolsep 4pt 
\begin{tabular}{l|c c c c c c|| c c c c c c}
\toprule
\multicolumn{1}{c|}{Metrics $\to$} & \multicolumn{6}{c||}{Mean Average Precision (MAP)} & \multicolumn{6}{c}{HITS @ 20} \\ \hline 
& \aids & \mutag & \fm & \fr& \mm & \mr & \aids & \mutag & \fm & \fr& \mm & \mr \\
\midrule \midrule 
\graphsim~\cite{graphsim} & 0.356 & 0.472 & 0.477 & 0.423 & 0.415 & 0.453 & 0.145 & 0.257 & 0.261 & 0.227 & 0.212 & 0.23 \\
\gotsim~\cite{gotsim} & 0.324 & 0.272 & 0.355 & 0.373 & 0.323 & 0.317 & 0.112 & 0.088 & 0.147 & 0.166 & 0.119 & 0.116\\
\simgnn~\cite{simgnn} & 0.341 & 0.283 & 0.473 & 0.341 & 0.298 & 0.379 & 0.138 & 0.087 & 0.235 & 0.155 & 0.111 & 0.160\\
\egsc~\cite{egsc} & 0.505 & 0.476 & 0.609 & 0.607 & 0.586 & 0.58 & 0.267 & 0.243 & 0.364 & 0.382 & 0.348 & 0.325\\
\hmn~\cite{zhang2021h2mn} & 0.267 & 0.276 & 0.436 & 0.412 & 0.312 & 0.243 & 0.076 & 0.084 & 0.200 & 0.189 & 0.119 & 0.069\\ 
\neuromatch~\cite{neuromatch} & 0.489 & 0.576 & 0.615 & 0.559 & 0.519 & 0.606 & 0.262 & 0.376 & 0.389 & 0.350 & 0.282 & 0.385\\
\greed~\cite{greed} & 0.472 & 0.567 & 0.558 & 0.512 & 0.546 & 0.528 & 0.245 & 0.371 & 0.316 & 0.287 & 0.311 & 0.277\\
\gen~\cite{gmn} & 0.557 & 0.605 & 0.661 & 0.575 & 0.539 & 0.631 & 0.321 & 0.429 & 0.448 & 0.368 & 0.292 & 0.391\\
\gmn~\cite{gmn} & 0.622 & \lfirst{0.710} & 0.730 & 0.662 & 0.655 & 0.708 & 0.397 & \lfirst{0.544} & 0.537 & 0.45 & 0.423 & 0.49\\ 
\isonode~\cite{RoyVCD2022IsoNet} & 0.659 & 0.697 & 0.729 & 0.68 & 0.708 & 0.738 & 0.438 & 0.509 & 0.525 & 0.475 & 0.493 & 0.532\\
\isoedge~\cite{RoyVCD2022IsoNet}  & \lfirst{0.690} & 0.706 & \lfirst{0.783} & \lfirst{0.722} & \lfirst{0.753} & \lfirst{0.774} & \lfirst{0.479} & 0.529 & \lfirst{0.613} & \lfirst{0.538} & \lfirst{0.571} & \lfirst{0.601}\\ \hline  
\ournodeshort & \second{0.825} & \second{0.851} & \second{0.888} & \second{0.855} & \second{0.838} & \second{0.874} & \second{0.672} & \second{0.732} & \second{0.797} & \second{0.737} & \second{0.702} & \second{0.755}\\
\ouredgeshort & \first{0.847} & \first{0.858} & \first{0.902} & \first{0.875} & \first{0.902} & \first{0.902} & \first{0.705} & \first{0.749} & \first{0.813} & \first{0.769} & \first{0.809} & \first{0.803}\\
 \hline \hline  
\end{tabular}}
\vspace{1mm}
\caption{Comparison of the two variants of \our\ (\ournode and \ouredge) against all the state-of-the-art graph retrieval methods, across all six datasets. Performance is measured in terms average precision (MAP) and mean HITS@20. In all cases, we used $60$\% training, $15$\% validation and $25$\% test sets.
The numbers highlighted with \fboxg{green} and \fboxy{yellow} indicate the best, second best method respectively, whereas the numbers with \fboxb{blue} indicate the best method among the baselines.  (MAP values for \ouredge across \fm, \mm\ and \mr\ were verified to be not exactly the same, but they match up to the third decimal place.)}
\vspace{-2mm}
\label{tab:main}
\end{table} 

\paragraph{Comparison with baselines} First, we compare \ournode\ and \ouredge\ against
all the baselines, across all datasets. 
In Table~\ref{tab:main}, we report the results. The key observations are as follows:
\textbf{(1)}~\ournode\ and \ouredge\ outperform all the baselines by   significant margins across all datasets. \ouredge consistently outperforms \ournode. This is because edge alignment allows us to compare the graph pairs more effectively than node alignment.  A similar effect was seen for \isoedge{} vs.\ \isonode~\cite{RoyVCD2022IsoNet}.
\textbf{(2)}~Among all state-of-the-art competitors, \isoedge\ performs the best followed by \isonode. Similar to us, they also use edge and node alignments respectively. 
However, \isonet\ does not perform any interaction between the graph pairs and the alignment is computed once only during the computation of $\Delta(G_c \given G_q)$. This results in modest performance compared to \our.
\textbf{(3)}~\gmn\ uses ``attention'' to estimate the alignment between graph pairs, which induces a non-injective mapping. Therefore, despite being an early interaction model, it is mostly outperformed by \isonet, which uses injective alignments.

\begin{table*}
\begin{minipage}{0.48\columnwidth}
\centering
\adjustbox{max width=.99\hsize}{
\begin{tabular}{l|cccccc}
\hline
& \aids & \mutag & \fm & \fr & \mm & \mr\\ \hline\hline
\multirow{2}{0.1cm}{\hspace{-0.5cm}\begin{sideways}{\textbf{Node}}\end{sideways}}\hspace{-4mm}\ldelim\{{2}{0.3mm}\hspace{2mm}
  Eager & 0.756 & 0.81 & 0.859 & 0.802 & 0.827 & 0.841 \\[1ex]
     Lazy & \first{0.825} & \first{0.851} & \first{0.888} & \first{0.855} & \first{0.838} & \first{0.874} \\ \hline\hline
  \multirow{2}{0.1cm}{\hspace{-0.5cm}\begin{sideways}{\textbf{Edge}}\end{sideways}}\hspace{-4mm}\ldelim\{{2}{0.3mm}\hspace{2mm}
   Eager & 0.795 & 0.805 & 0.883 & 0.812 & 0.862 & 0.886 \\[1ex] 
 Lazy  & \first{0.847} & \first{0.858} & \first{0.902} & \first{0.875} & \first{0.902} & \first{0.902} \\ \hline\hline
 \end{tabular}}
\caption{\small Lazy multi-round vs.\ eager multi-layer. First (Last) two rows report MAP for \ournode\ (\ouredge). \fboxg{Green} shows the best method.}
\label{tab:multi}
\end{minipage} \hspace{0.02\columnwidth}
\begin{minipage}{0.51\columnwidth}
\centering
\adjustbox{max width=.99\hsize}{
\begin{tabular}{l|cccccc}
\hline
& \aids & \mutag & \fm & \fr & \mm & \mr \\ \hline\hline
\multirow{2}{0.1cm}{\hspace{-0.65cm}\begin{sideways}{\textbf{Lazy}}\end{sideways}}
\hspace{-6mm}\ldelim\{{2}{0.3mm}\hspace{2mm}
\post & 0.776 & 0.829 & 0.851 & 0.819 & \first{0.844} & 0.84 \\[1ex] 
\ournodeshort & \first{0.825} & \first{0.851} & \first{0.888} & \first{0.855} & {0.838} & \first{0.874}  \\ \hline\hline
\multirow{2}{0.1cm}{\hspace{-0.65cm}\begin{sideways}{\textbf{Eager}}\end{sideways}}
\hspace{-6mm}\ldelim\{{2}{0.3mm}\hspace{2mm}
\post & 0.668 & 0.783 & 0.821 & 0.752 & 0.753 & 0.794 \\[1ex] 
\ournodeshort & \first{0.756} & \first{0.81} & \first{0.859} & \first{0.802} & \first{0.827} & \first{0.841}  \\ \hline\hline
 \end{tabular}}
 \caption{\small Node partner vs. node pair partner interaction.
 First (Last) two rows report MAP for multi-round (multi-layer) update.  \fboxg{Green} shows the best method. }
\label{tab:cross}
\end{minipage}
  \vspace{-1mm}
\end{table*}

\paragraph{Lazy vs.\ eager updates} In lazy multi-round updates, the alignment matrices remain unchanged across all propagation layers and are updated only after the GNN completes its $K$-layer message propagations. To evaluate its effectiveness,  we compare it against   the eager multi-\emph{layer} update (described at the end of Section~\ref{subsec:our_node_design}), where
the GNN executes its $K$-layer message propagations only once; the alignment map is updated across $K$ layers; and, the alignment at $k$th layer is used to compute the embeddings at $(k+1)$th layer. 
In Table~\ref{tab:multi}, we compare the performance in terms MAP, which shows that lazy multi-round updates significantly outperform multi-layer updates. 

\paragraph{Node partner vs. node-pair partner  interaction}  To understand the benefits of node-pair partner interaction, we contrast \ournode\ against another variant of our method, which performs \emph{node partner} interaction  rather than node pair partner interaction, similar to Eq.~\eqref{eq:gmnupdate}. For lazy multi-round updates, we compute the embeddings as follows:
\begin{align}
\hq_{t+1,k+1}(u) &= \textstyle\comb_{\theta} (
\hq_{t+1, k}(u), \; \sum_{v\in\nbr(u)} \msg_{\theta}(\hq_{t,k}(u), \hq_{t,k}(v)), \;
 \sum_{u'\in V_c}\Pb_t[u,u']\hc_{t,k}(u')  ) \nn
\end{align}
For eager multi-layer updates, we compute the embeddings as:
\begin{align}
\hq_{k+1}(u) &= \textstyle\comb_{\theta}(
\hq_{k}(u), \; \sum_{v\in\nbr(u)} \msg_{\theta}(\hq_{k}(u), \hq_{k}(v)), \;
 \sum_{u'\in V_c}\Pb_k[u,u']\hc_{k}(u') ) \nn\label{eq:post-main-paper-2}
\end{align}
Table~\ref{tab:cross} 
summarizes the results, which shows that \ournode\ (node partner pair) performs significantly better than  \post\ for both multi-round lazy updates (top-two rows) and multi-layer eager updates (bottom tow rows).

\begin{figure}[!t]
    \centering
    \vspace{-4mm} 
\subfloat{\includegraphics[width=.24\linewidth]{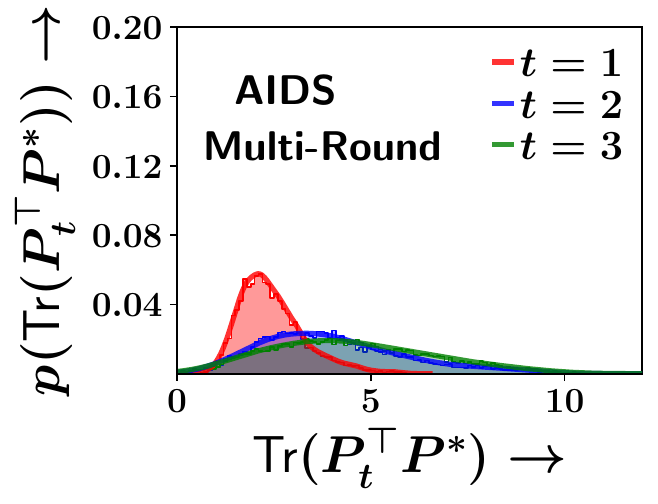}}
\subfloat{\includegraphics[width=.24\linewidth]{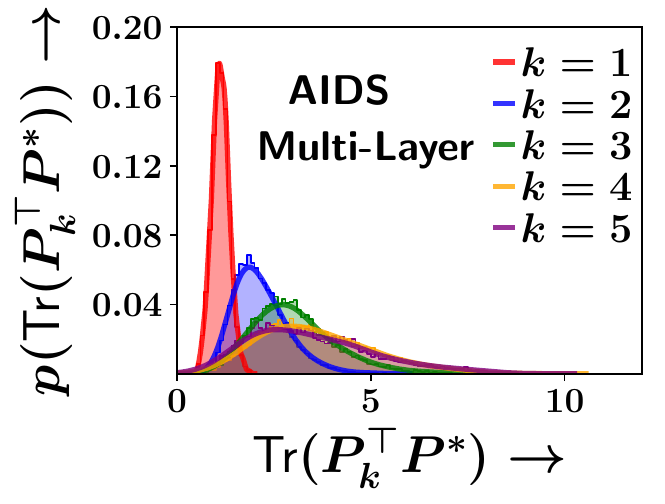}}\hspace{3mm}
\subfloat{\includegraphics[width=.24\linewidth]{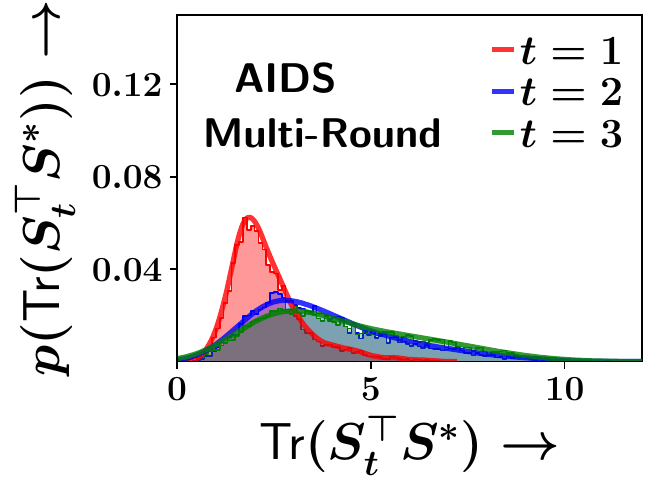}} 
\subfloat{\includegraphics[width=.24\linewidth]{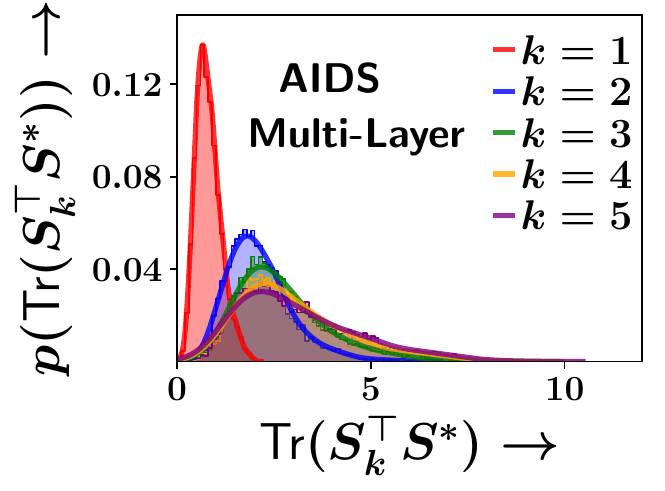}}
        \caption{Empirical probability density of  similarity between the estimated alignments and the true alignments $\Pb^*,\Sb^*$ for both multi-round and multi-layer update strategies across different stages of updates ($t$ for multi-round and $k$ for multi-layer), for \aids.  
        %
        Similarity is measured using  $p(\text{Tr}(\Pb_t ^{\top}\Pb^*)),  p(\text{Tr}(\Sb_t ^{\top}\Sb^*))$ for multi-round lazy updates and $p(\text{Tr}(\Pb_k ^{\top}\Pb^*)),  p(\text{Tr}(\Sb_k ^{\top}\Sb^*))$ for multi-layer eager updates.}
        \label{fig:hist}
\end{figure}

\paragraph{Quality of injective alignments}
Next we compare between multi-round and multi-layer update strategies in terms of their ability to refine the alignment matrices, as the number of updates of these matrices increases.  For multi-round (layer) updates, we instrument the alignments $\Pb_t$ and $\Sb_t$ ($\Pb_k$ and $\Sb_k$) for different rounds $t\in [T]$ (layers $k\in [K]$). Specifically, we look into the distribution of the similarity between the learned alignments $\Pb_t,\Sb_t$ and the correct alignments $\Pb^*,\Sb^*$ (using combinatorial routine), measured using the inner products $\text{Tr}(\Pb_t ^{\top}\Pb^*)$ and $\text{Tr}(\Sb_t ^{\top}\Sb^*)$ for different $t$. Similarly, we compute
$\text{Tr}(\Pb_k ^{\top}\Pb^*)$ and $\text{Tr}(\Sb_k ^{\top}\Sb^*)$ for different $k\in[K]$.
Figure~\ref{fig:hist} summarizes the results, which shows that \textbf{(1)}~as $t$ or $k$ increases, the learned alignments become closer to the gold alignments; \textbf{(2)}~multi-round updates refine the alignments approximately twice as faster  
than the multi-layer variant.  The distribution of $\text{Tr}(\Pb_t ^{\top}\Pb^*)$ at $t=1$ in multi-round strategy is almost always close to $\text{Tr}(\Pb_k ^{\top}\Pb^*)$ for $k=2$.
Note that, our aligner networks learn to refine the $\Pb_t$ and $\Sb_t$ through end-to-end training, without using any form of supervision from true alignments or the gradient computed in Eq.~\eqref{eq:PGWupdate}.

\begin{figure}[h]
    \centering
\subfloat[\vspace{-3mm} Node, \aids]{\includegraphics[width=.3\linewidth]{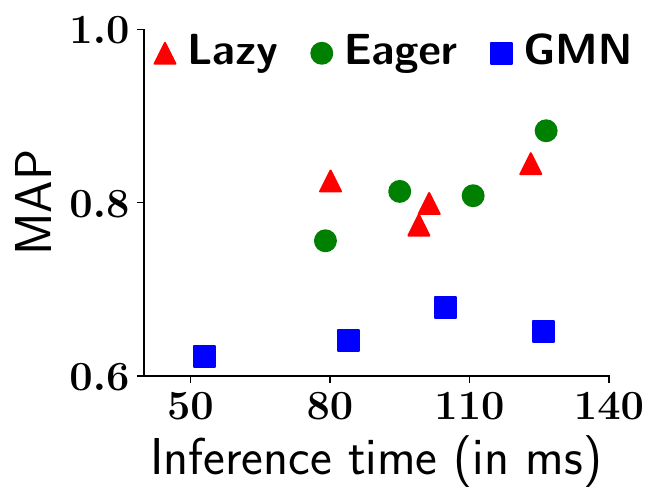}} \hspace{1cm}
\subfloat[Edge, \aids]{\includegraphics[width=.3\linewidth]{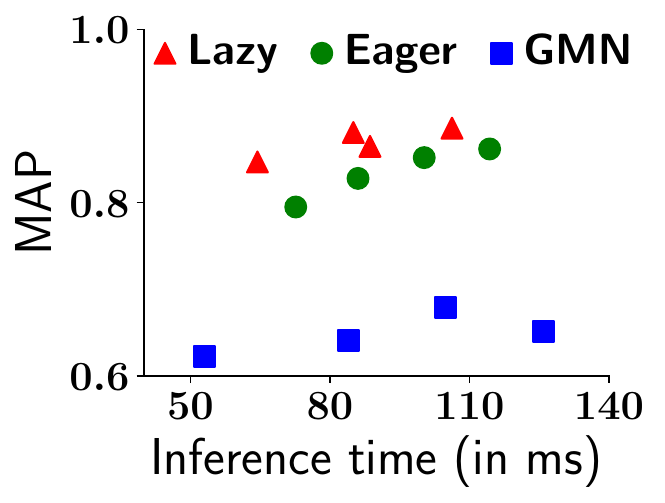}}
 \caption{Trade-off between MAP and inference time (batch size=128).}
 \label{fig:tradeoff}
\end{figure}


\paragraph{Accuracy-inference time trade-off} 
Here, we analyze the accuracy and inference time trade-off. We vary $T$ and $K$ for \our's lazy multi-round variant, and vary $K$ for \our's eager multi-layer variant and  for  \gmn.  Figure~\ref{fig:tradeoff} summarizes the results. Notably, the eager multi-layer variant achieves the highest accuracy for $K=8$ on the \aids dataset, despite the known issue of oversmoothing in GNNs for large $K$. This unexpected result may be due to our message passing components, which involve terms like $\sum_{u'}\Pb[u,u'] \hb(u')$, effectively acting as a convolution between alignment scores and embedding vectors. This likely enables $\Pb$ to function as a filter, countering the oversmoothing effect.

\section{Conclusion}
\label{sec:End}
We introduce \our{} as an early-interaction network for estimating subgraph isomorphism. \our{} learns to identify explicit alignments between query and corpus graphs despite having access to only pairwise preferences and not explicit alignments during training.  We design a graph neural network (GNN) that uses an alignment estimate to propagate messages, then uses the GNN's output representations to refine the alignment.  Experiments across several datasets confirm that alignment refinement is achieved over several rounds.  Design choices such as using node-pair partner interaction (instead of node partner) and lazy updates (over eager) boost the performance of our architecture, making it the state-of-the-art in subgraph isomorphism based subgraph retrieval. We also demonstrate the accuracy v/s inference time trade offs for \our, which show how different knobs can be tuned to utilize our models under regimes with varied time constraints.

This study can be extended to graph retrieval problems which use different graph similarity measures, such as maximum common subgraph or graph edit distance. Extracting information from node-pairs is effective and can be widely used to improve GNNs working on multiple graphs at once.
\section*{Acknowledgements}
Indradyumna acknowledges Qualcomm Innovation Fellowship, Abir  and  Soumen acknowledge grants from Amazon, Google, IBM and SERB.

 
\bibliography{voila,refs}
\bibliographystyle{abbrvnat}

\newpage
\appendix
\onecolumn

\begin{center}
    {\Large \bf \ztitle \\ (Appendix)}
\end{center}

\section{Limitations}
\label{app:lim}
We find two limitations of our method each of which could form the basis of detailed future studies.
\begin{enumerate}
    \item Retrieval systems greatly benefit from the similarity function being hashable. This can improve the inference time multi-fold while losing very little, if at all any, performance, making the approach ready for production environments working under tight time constraints. The design of a hash function for an early interaction network like ours is unknown and seemingly difficult. In fact, such a hashing procedure is not known even for predecessors like \isoedge~or \gmn, and this is an exciting future direction.

    \item Our approach does not explicitly differentiate between nodes or edges that may belong to different classes. This can be counterproductive when there exist constraints that prevent the alignment of two nodes or edges with different labels. While the network is designed to process node and edge features, it might not be enough to rule out alignments that violate the said constraint. Such constraints could also exist for node-pairs, such as in knowledge graphs with hierarchical relationships between entity types, and are not taken into account by our model. Extending our work to handle such restrictions is an interesting problem to consider.
\end{enumerate}

\section{Related work}
\label{app:related}
In this section, we discuss different streams of work that are related to and have influenced the study.
\subsection{Graph Representation Learning}
Graph neural networks (GNN)~\cite{gilmer2017neural, gmn, li2015gated, kipf2016semi, xu2018powerful, velivckovic2017graph} have emerged as a widely applicable approach for graph representation learning. A graph neural network computes the embedding of a node by aggregating the representations of its neighbors across $K$ steps of message passing, effectively combining information from $K$-hop neighbors. GNNs were first used for graph similarity computation by~\citet{gmn}, who enriched the architecture with attention to predict isomorphism between two graphs. Attention acts as a mechanism to transfer information from the representation of one graph to that of the other, thus boosting the performance of the approach. \citet{chen2022structure} enriched the representation of graphs by capturing the subgraph around a node effectively through a structure aware transformer architecture.

\subsection{Differentiable combinatorial solvers}
We utilize a differentiable gadget to compute an injective alignment, which is a doubly stochastic matrix. The differentiability is crucial to the training procedure as it enables us to backpropagate through the alignments. The $\operatorname{GumbelSinkhorn}$ operator, which performs alternating normalizations across rows and columns, was first proposed by~\citet{sinkhorn1967concerning} and later used for the Optimal Transport problem by~\citet{cuturi}. Other methods to achieve differentiability include adding random noise to the inputs to discrete solvers~\cite{berthet2020learning} and designing probabilistic loss functions~\cite{karalias2020erdos}. A compilation of such approaches towards constrained optimization on graphs through neural techniques is presented in~\cite{kotary2021end}.

\subsection{Graph Similarity Computation and Retrieval}
Several different underlying measures have been proposed for graph similarity computation, including full graph isomorphism~\cite{gmn}, subgraph isomorphism~\cite{neuromatch, RoyVCD2022IsoNet}, graph edit distance (GED)~\cite{graphsim, gotsim, ged3, ged2, ged1} and maximum common subgraph (MCS)~\cite{graphsim, gotsim,RoyCD2022McsNet}. \citet{graphsim} proposed~\graphsim towards the GED and MCS problems, using convolutional neural network based scoring on top of graph similarity matrices. GOTSim~\cite{gotsim} explicitly computes the alignment between the two graphs by studying the optimal transformation cost. GraphSim~\cite{graphsim} utilizes both graph-level and node-level signals to compute a graph similarity score. NeuroMatch~\cite{neuromatch} evaluates, for each node pair across the two graphs, if the neighborhood of one node is contained in the neighborhood of another using order embeddings~\cite{McFee2009PartialOE}. GREED~\cite{greed} proposed a Siamese graph isomorphism network, a late interaction model to tackle the GED problem and provided supporting theoretical guarantees. \citet{zhang2021h2mn} propose an early interaction model, using hypergraphs to learn higher order node similarity. Each hypergraph convolution contains a subgraph matching module to learn cross graph similarity. \citet{egsc} trained a slower attention-based network on multi-level features from a GNN and distilled its knowledge into a faster student model. \citet{RoyVCD2022IsoNet} used the $\operatorname{GumbelSinkhorn}$ operator as a differentiable gadget to compute alignments in a backpropagation-friendly fashion and also demonstrated the utility of computing alignments for edges instead of nodes.

\section{Broader Impact}
\label{app:impact}
This work can be directly applied to numerous practical applications, such as drug discovery and circuit design, which are enormously beneficial for the society and continue to garner interest from researchers and practitioners worldwide. The ideas introduced in this paper have benefitted from and can benefit the information retrieval community as well, beyond the domain of graphs. However, malicious parties could use this technology for deceitful purposes, such as identifying and targeting specific social circles on online social networks (which can be represented as graphs). Such pros and cons are characteristic of every scientific study and the authors consider the positives to far outweigh the negatives.

\section{Network architecture of different components of \our}
\label{app:model}
\our~models consist of three components - an encoder, a message-passing network and a node/edge aligner. We provide details about each of these components below. For convenience, we represent a linear layer with input dimension $a$ and output dimension $b$ as $\linear(a,b)$ and a linear-ReLU-linear network with $\linear(a,b),\ \linear(b,c)$ layers with ReLU activation in the middle as $\lrl(a,b,c)$.

\subsection{Encoder}
The encoder transforms input node/edge features before they are fed into the message-passing network. For models centred around node alignment like \ournode, the encoder refers to $\init_{\theta}$ and is implemented as a $\linear(1,10)$ layer. The edge vectors are not encoded and passed as-is down to the message-passing network. For edge-based models like \ouredge, the encoder refers to both $\init_{\theta,\text{node}}$ and $\init_{\theta,\text{edge}}$, which are implemented as $\linear(1,10)$ and $\linear(1,20)$ layers respectively.

\subsection{GNN}
 Within the message-passing framework, we use node embeddings of size $\dim_h=10$ and edge embeddings of size $\dim_m=20$. We specify each component of the GNN below.
\begin{itemize}
    \item $\inter_\theta$ combines the representation of the current node/edge ($\hb_{\bullet}$) with that from the other graph, which are together fed to the network by concatenation. For node-based and edge-based models, it is implemented as $\lrl(20,20,10)$ and $\lrl(40,40,20)$ networks respectively. In particular, we ensure that the input dimension is twice the size of the output dimension, which in turn equals the intermediate embedding dimension $\dim(\zb)$.
    \item $\msg_\theta$ is used to compute messages by combining intermediate embeddings $\zb_{\bullet}$ of nodes across an edge with the representation of that edge. For node-based models, the edge vector is a fixed vector of size $1$ while the intermediate node embeddings $\zb_{\bullet}$ are  vectors of dimension $10$, resulting in the network being a $\linear(21, 20)$ layer. For edge-based models, the edge embedding is the $\bm{m}$ vector of size $20$ which requires $\msg_\theta$ to be a $\linear(40, 20)$ layer. 
    Note that the message-passing network is applied twice, once to the ordered pair $(u,v)$ and then to $(v,u)$ and the outputs thus obtained are added up. This is to ensure node order invariance for undirected edges by design.
    \item $\comb_\theta$ combines the representation of a node $\zb_\bullet$ with aggregated messages received by it from all its neighbors. It is modelled as a $\gru$ where the node representation (of size $10$) is the initial hidden state and the aggregated message vector (of size $20$) is the only element of an input sequence which updates the hidden state to give us the final node embedding $\hb_\bullet$.
\end{itemize} 

\subsection{Node aligner}
\label{subsec:node-aligner}
The node aligner takes as input two sets of node vectors $\Hq\in\RR^{n \times 10}$ and $\Hc\in\RR^{n \times 10}$ representing $G_q$ and $G_c$ respectively. $n$ refers to the number of nodes in the corpus graph (the query graph is padded to meet this node count). We use $\operatorname{LRL}_{\phi}$ as a  $\lrl(10, 16, 16)$ network (refer Eq.~\ref{eq:Pt+1}).

\subsection{Edge aligner}
The design of the edge aligner is similar to the node aligner described above in Section~\ref{subsec:node-aligner}, except that its inputs are sets of edge vectors $\Mq\in\RR^{e \times 20}$ and $\Mc\in\RR^{e \times 20}$. $e$ refers to the number of edges in the corpus graph (the query graph is padded to meet this edge count). We use $\text{LRL}_{\phi}$ as a  $\lrl(20, 16, 16)$ network (refer Eq.~\ref{eq:St+1}).

\subsection{$\operatorname{GumbelSinkhorn}$ operator}
\label{subsec:gumbel-sinkhorn}

The $\operatorname{GumbelSinkhorn}$ operator consists of the following operations -
\begin{align}
  &  \bm{D}_0 = \operatorname{exp}(\bm{D}_{\text{in}} / \tau)\\
   & \bm{D}_{t+1} = \operatorname{RowNorm}\left(\operatorname{ColumnNorm}(\bm{D}_t)\right)\label{eq:Dt-Dt+1}\\
  &  \bm{D}_{\text{out}} = \lim_{t\to\infty}\bm{D}_t
\end{align}
The matrix $\bm{D}_{\text{out}}$ obtained after this set of operations will be a doubly-stochastic matrix. The input $\bm{D}_{\text{in}}$ in our case is the matrix containing the dot product of the node/edge embeddings of the query and corpus graphs respectively. $\tau$ represents the temperature and is fixed to $0.1$ in all our experiments.

\paragraph{Theorem} Equation~\ref{eq:Pt+1} results in a permutation matrix that is row-equivariant (column-) to the shuffling of nodes in $G_q$ ($G_c$).

\paragraph{Proof} To prove the equivariance of Eq.~\ref{eq:Pt+1}, we need to show that given a shuffling (permutation) of query nodes $Z\in \Pi_{n}$ which modifies the node embedding matrix to $Z\dot \Hb_{t,K}^{(q)}$, the resulting output of said equation would change to $Z\Pb_t$. Below, we consider any matrices with $Z$ in the suffix as being an intermediate expression in the computation of $\operatorname{NodeAlignerRefinement}_\phi(Z\Hb_{t,K}^{(q)}, \Hb_{t,K}^{(c)})$.

It is easy to observe that the operators $\operatorname{LRL}_\phi$ (a linear-ReLU-linear network applied to a matrix), $\operatorname{RowNorm}$, $\operatorname{ColumnNorm}$ and element-wise exponentiation ($\operatorname{exp}$), division are all permutation-equivariant since a shuffling of the vectors fed into these will trivially result in the output vectors getting shuffled in the same order. Thus, we get the following sequence of operations
\begin{equation}
\bm{D}_\text{in,$Z$}=\operatorname{LRL}_\phi(Z\Hb_{t,K}^{(q)})\operatorname{LRL}_\phi(\Hb_{t,K}^{(c)})^\top = Z\cdot\operatorname{LRL}_\phi(\Hb_{t,K}^{(q)})\operatorname{LRL}_\phi(\Hb_{t,K}^{(c)})^\top\bm{D}_\text{in}=Z\bm{D}_\text{in}
\end{equation}
$\bm{D}_{0,Z}$ equals $\operatorname{exp}(\bm{D}_\text{in,$Z$} / \tau)$, which according to above equation would lead to $\bm{D}_{0,Z} = Z\bm{D}_{0}$. We can then inductively show using Eq.~\ref{eq:Dt-Dt+1} and the equivariance of row/column normalization, assuming the following holds till $t$, that 
\begin{align}
&\bm{D}_{t+1,Z}=\operatorname{RowNorm}\left(\operatorname{ColumnNorm}(\bm{D}_{t,Z})\right)=\operatorname{RowNorm}\left(\operatorname{ColumnNorm}(Z\bm{D}_{t})\right)\\
&=\operatorname{RowNorm}\left(Z\cdot\operatorname{ColumnNorm}(\bm{D}_{t})\right)=Z\cdot\operatorname{RowNorm}\left(\operatorname{ColumnNorm}(\bm{D}_{t})\right)=Z\bm{D}_{t+1}
\end{align}
The above equivariance would also hold in the limit, resulting in the doubly stochastic matrix $\bm{D}_{\text{out},Z} = Z\bm{D}_\text{out}$, which concludes the proof.
\hspace{7cm}
$\blacksquare$
A similar proof can be followed to show column equivariance for a shuffling in the corpus nodes.

\section{Variants of our models and GMN, used in the experiments}
\label{app:variants}


\subsection{Multi-round refinement of \ournode for the corpus graph}
\label{app:multi-round-for-corpus}
\begin{itemize}
    \item Initialize:
\begin{align}
    \hc_{0} (u') =  \init_{\theta}(\feat(u') ), \label{eq:init-node-multiround-corpus}
\end{align}
    \item Update the GNN embeddings as follows:
\begin{align}
    &\zc_{t+1,k}(u') = \textstyle \inter_{\theta}\Big(\hc_{t+1,k}(u'), \sum_{u\in V_q} \hq_{t,k}(u) \Pb_{t}^\top[u',u]\Big),   \label{eq:inter-1-corpus}\\
   & \hctk{t+1}{k+1}(u')  =  \textstyle\comb_{\theta}\Big(\zc_{t+1, k} (u'),  
    \sum_{v'\in \nbr(u')} \msg_{\theta}(\zc_{t+1, k} (u'), \zc_{t+1, k} (v')) \Big)     \label{eq:node-emb-n1-corpus}
\end{align}
\end{itemize}
 
\subsection{Eager update  for \ouredge}
\begin{itemize}
    \item Initialize:
        \begin{align}
            \hq_{0} (u) =  \init_{\theta, \text{node}}(\feat(u) ), \label{eq:init-edge-node-multilayer}\\
            \mq_{0} (e) =  \init_{\theta, \text{edge}}(\feat(e) ), \label{eq:init-edge-edge-multilayer}
        \end{align}
    \item The edge alignment is updated across layers. $\Sb_{0}$ is set to a matrix of zeros. For $k>0$, the following equation is used:
        \begin{align}
        \Sb_{k}&  = \operatorname{EdgeAlignerRefinement}_{\phi}\big(\Mq_{k},\Mc_{k} \big) \\
            & \begin{aligned}
              &  = \operatorname{GumbelSinkhorn}\left( \operatorname{LRL}_\phi(\Mq_{k})
              \operatorname{LRL}_\phi(\Mc_{k}) ^\top \right) \label{eq:St+1-multilayer}
            \end{aligned}
        \end{align}

    \item We update the GNN node and edge embeddings as follows:
         \begin{align}
            &\zq_{k}(e) = \textstyle \inter_{\theta}\Big(\text{{$\mq_{k}(e),$}} \sum_{e'\in E_c} \mc_{k}(e') \Sb_{k}[e,e']\Big), \label{eq:inter-edge-multilayer}
    \\
            &  \hq_{k+1}(u)   =  \textstyle\comb_{\theta}\Big(\hq_{k}(u),  
            \sum_{a\in \nbr(u)} \msg_{\theta}(\hq_{k}(u), \hq_{k}(a), \zq_{k}((u,a))) \Big)  \label{eq:node-emb-edge-multilayer}
 \\
   &         \mq_{k+1}((u,v)) =\msg_{\theta}(\hq_{k+1}(u), \hq_{k+1}(v), \zq_{k}((u,v)))
        \end{align}
\end{itemize}

\subsection{\uonly variant of \ournode}
\label{app:uonly-equations}
Here, we update node embeddings as follows:
\begin{align}
    \hqtk{t+1}{k+1}(u) &=  \textstyle\comb_{\theta}\Big(\zq_{t+1, k} (u), \sum_{v\in \nbr(u)} \underbrace{\msg_{\theta}(\hq_{t+1, k} (u), \hq_{t+1, k} (v))}_{\text{ $\zb$ is replaced with $\hb$}} \Big)
\end{align}
Here, $\zq_{t+1, k} (u)$ is computed as Eq.~\eqref{eq:inter-1}, where $\text{inter}_{\theta}$ is an MLP network.  
In contrast to Eq.~\eqref{eq:node-emb-n1}, here, $\zq_{t+1, k} (u), \zq_{t+1, k} (v)$ are not fed into the message passing layer. Hence, in the message passing layer, we do not capture the signals from the partners of $u$ and $v$ in $G_c$.
Only signals from partners of $u$ are captured through $\zq_{t+1, k} (u)$ in the first argument.

\subsection{\monly variant of \ournode}
\label{app:monly-equations}
We change the GNN update equation as follows:
\begin{align}
    \hqtk{t+1}{k+1}(u) &=  \textstyle\comb_{\theta}\Big(\underbrace{\hq_{t+1, k} (u)}_{\text{$\zb$ is replaced with $\hb$}}, \sum_{v\in \nbr(u)} \msg_{\theta}(\zq_{t+1, k} (u), \zq_{t+1, k} (v)) \Big)
\end{align}
Node pair partner interaction takes place because, we feed $\zb$ from Eq.~\eqref{eq:inter-1} into the message passing layer. However, we use $\hb$ in the first argument, instead of $\zb$.

\newpage
\section{Additional details about experimental setup}
\label{app:details}

\subsection{Datasets}
\label{app:datasets}
We use six datasets from the TUDatasets collection~\cite{Morris+2020} for benchmarking our methods with respect to existing baselines. \citet{neuromatch} devised a method to sample query and corpus graphs from the graphs present in these datasets to create their training data. We adopt it for the task of subgraph matching. In particular, we choose a node $u \in G$ as the center of a Breadth First Search (BFS) and run the algorithm till $|V|$ nodes are traversed, where the range of $|V|$ is listed in Table~\ref{tab:dataset-stats} (refer to the Min and Max columns for $|V_q|$ and $|V_c$). This process is independently performed for the query and corpus splits (with different ranges for graph size) to obtain $300$ query graphs and $800$ corpus graphs. The set of query graphs is split into train, validation and test splits of $180\ (60\%)$, $45\ (15\%)$ and $75\ (25\%)$ graphs respectively. Ground truth labels are computed for each query-corpus graph pair using the VF2 algorithm~\cite{cordella2004sub, hagberg2008exploring, neuromatch} implemented in the Networkx library. Various statistics about the datasets are listed in Table~\ref{tab:dataset-stats}. $\operatorname{pairs}(y)$ denotes the number of pairs in the dataset with gold label $y$, where $y\in\{0,1\}$.

\begin{table}[!ht]
\centering
\adjustbox{max width=\hsize}{\tabcolsep 2pt  \small
\begin{tabular}{l|ccccccccccc}
\hline
 & Mean $|V_q|$ & Min $|V_q|$ & Max $|V_q|$ & Mean $|E_q|$ & Mean $|V_c|$ & Min $|V_c|$ & Max $|V_c|$ & Mean $|E_c|$ & $\operatorname{pairs(1)}$ & $\operatorname{pairs(0)}$ &  $\frac{\operatorname{pairs(1)}}{\operatorname{pairs(0)}}$ \\ \hline\hline
\aids & 11.61 & 7 & 15 & 11.25 & 18.50 & 17 & 20 & 18.87 & 41001 & 198999 & 0.2118 \\
\mutag & 12.91 & 6 & 15 & 13.27 & 18.41 & 17 & 20 & 19.89 & 42495 & 197505 & 0.2209 \\
\fm & 11.73 & 6 & 15 & 11.35 & 18.30 & 17 & 20 & 18.81 & 40516 & 199484 & 0.2085 \\
\fr & 11.81 & 6 & 15 & 11.39 & 18.32 & 17 & 20 & 18.79 & 39829 & 200171 & 0.2043 \\
\mm & 11.80 & 6 & 15 & 11.37 & 18.36 & 17 & 20 & 18.79 & 40069 & 199931 & 0.2056 \\
\mr & 11.87 & 6 & 15 & 11.49 & 18.32 & 17 & 20 & 18.78 & 40982 & 199018 & 0.2119 \\\hline\hline
 \end{tabular}}
\caption{\small Statistics for the $6$ datasets borrowed from the TUDatasets collection~\cite{Morris+2020}}
\label{tab:dataset-stats}
\end{table}

\subsection{Baselines}
\textbf{\graphsim, \gotsim, \simgnn, \neuromatch, \gen, \gmn, \isonode, \isoedge}: We utilized the code from official implementation of ~\cite{RoyVCD2022IsoNet} \footnote{\url{https://github.com/Indradyumna/ISONET/}}. Some \textit{for loops} were vectorized to improve the running time of \gmn.\\
\textbf{\egsc}: The official implementation \footnote{\url{https://github.com/canqin001/Efficient_Graph_Similarity_Computation}} is refactored and integrated into our code.\\
\textbf{\hmn}: We use the official code from \footnote{\url{https://github.com/cszhangzhen/H2MN}}.\\
\textbf{\greed}: We use the official code from \footnote{\url{https://github.com/idea-iitd/greed}}. The model is adapted from the graph edit distance (GED) task to the subgraph isomorphism task, using a hinge scoring layer.

The number of parameters involved in all models (our methods and baselines) are reported in Table~\ref{tab:param-count}.
\begin{table}[h]
\centering
\maxsizebox{0.99\hsize}{!}{  \tabcolsep 7pt 
\begin{tabular}{l|c}
\toprule
& Number of parameters \\
\midrule \midrule 
\graphsim~\cite{graphsim} & 3909\\
\gotsim~\cite{gotsim} & 304\\
\simgnn~\cite{simgnn} & 1671\\
\egsc~\cite{egsc} & 3948\\
\hmn~\cite{zhang2021h2mn} & 2974\\
\neuromatch~\cite{neuromatch} & 3463\\
\greed~\cite{greed} & 1840\\
\gen~\cite{gmn} & 1750\\
\gmn~\cite{gmn} & 2050\\
\isonode~\cite{RoyVCD2022IsoNet} & 1868\\
\isoedge~\cite{RoyVCD2022IsoNet} & 2028\\
\ournode & 2498\\
\ouredge & 4908\\
\hline \hline
 \end{tabular}}
 \vspace{1mm}
\caption{Number of parameters for all models used in comparison}
\label{tab:param-count}
\end{table}

\subsection{Calculation of Metrics: Mean Average Precision (MAP), HITS@K, Precision@K and Mean Reciprocal Rank (MRR)}

Given a ranked list of corpus graphs $C=\set{G_c}$ for a test query $G_q$, sorted in the decreasing order of $\Delta_{\theta, \phi}(G_c|G_{q})$, let us assume that the $c_+^\text{th}$ relevant graph is placed at position $\text{pos}(c_+) \in \set{1,...,|C|}$ in the ranked list.  Then Average Precision (AP) is computed as:
\begin{align}
    \text{AP}(q) = \frac{1}{|C_{q+}|}\sum_{c_+\in [|C_{q+}|]} \frac{c_+}{ \text{pos}(c_+)}
\end{align}
Mean average precision is defined as $\sum_{q\in Q}\text{AP}(q)/|Q|$.

Precision@$K(q) = \frac{1}{K} \,{\text{\# relevant graphs corresponding to $G_q$ till rank $K$}}$. Finally we report the mean of Precision@$K(q)$ across queries.

Reciprocal rank or RR$(q)$ is the inverse of the rank of the topmost relevant corpus graph corresponding to $G_q$ in the ranked list. Mean reciprocal rank (MRR) is average of RR$(q)$ across queries.

 HITS@$K$ for a query $G_q$ is defined as the fraction of positively labeled corpus graphs that appear before the $K^\text{th}$ negatively labeled corpus graph. Finally, we report  the average of HITS@$K$ across queries.
 

Note that HITS@K is a significantly aggressive metric compared to Precision@K and MRR, as can be seen in Tables~\ref{tab:main-app-map-hits} and~\ref{tab:main-app-mrr-precision}.

\subsection{Details about hyperparameters}
All models were trained using early stopping with MAP score on the validation split as a stopping criterion. For early stopping, we used a patience of $50$ with a tolerance of $10^{-4}$. We used the Adam optimizer with the learning rate as $10^{-3}$ and the weight decay parameter as $5\cdot10^{-4}$. We set batch size to $128$ and maximum number of epochs to $1000$.

\paragraph{Seed Selection and Reproducibility} Five integer seeds were chosen uniformly at random from the range $[0, 10^4]$ resulting in the set $\{1704, 4929, 7366, 7474, 7762\}$. \ournode, \gmn~and \isoedge~were trained on each of these $5$ seeds for all $6$ datasets. Note that these seeds do not control the training-dev-test splits but only control the initialization. Since the overall problem is non-convex, in principle, one should choose the best initial conditions leading to local minima. Hence, for all models, we choose the 
best seed, based on validation MAP score, is shown in Table~\ref{tab:seeds}.

\begin{table}[!h]
\centering
\adjustbox{max width=\hsize}{\tabcolsep 7pt 
\begin{tabular}{l|c c c c c c}
\toprule
& \aids & \mutag & \fm & \fr& \mm & \mr  \\
\midrule \midrule 
\graphsim~\cite{graphsim} & 7762 & 4929 & 7762 & 7366 & 4929 & 7474\\
\gotsim~\cite{gotsim} & 7762 & 7366 & 1704 & 7762 & 1704 & 7366\\
\simgnn~\cite{simgnn} & 7762 & 7474 & 1704 & 4929 & 4929 & 7762\\
\egsc~\cite{egsc} & 4929 & 1704 & 7762 & 4929 & 4929 & 7366\\
\hmn~\cite{zhang2021h2mn} & 7762 & 4929 & 7366 & 1704 & 4929 & 7474\\
\neuromatch~\cite{neuromatch} & 7366 & 4929 & 7762 & 7762 & 1704 & 7366\\
\greed~\cite{greed} & 7762 & 1704 & 1704 & 7474 & 1704 & 1704\\
\gen~\cite{gmn} & 1704 & 4929 & 7474 & 7762 & 1704 & 1704\\
\gmn~\cite{gmn} & 7366 & 4929 & 7366 & 7474 & 7474 & 7366 \\
\isonode~\cite{RoyVCD2022IsoNet} & 7474 & 7474 & 7474 & 1704 & 4929 & 1704\\\hline
\isoedge~\cite{RoyVCD2022IsoNet} & 7474 & 7474 & 7474 & 1704 & 4929 & 1704 \\ 
\gmn~\cite{gmn} & 7366 & 4929 & 7366 & 7474 & 7474 & 7366 \\ 
\ournode & 7762 & 7762 & 7474 & 7762 & 7762 & 7366  \\ 
\hline \hline
 \end{tabular}}
 \vspace{1mm}
\caption{Best seeds for all models. For \isoedge, \gmn~and \ournode, these are computed based on MAP score on the validation split at convergence. For other models, the identification occurs after $10$ epochs of training.}
\label{tab:seeds}
\end{table}

\ouredge and all ablations on top of \ournode were trained using the best seeds for \ournode (as in Tables~\ref{tab:multi}, \ref{tab:cross} and \ref{tab:cross-our}). Ablations of \gmn~were trained with the best \gmn~seeds.

For baselines excluding \isoedge, models were trained on all $5$ seeds for $10$ epochs and the MAP scores on the validation split were considered. Full training with early stopping was resumed only for the best seed per dataset. This approach was adopted to reduce the computational requirements for benchmarking.

\paragraph{Margin Selection} For \graphsim, \gotsim, \simgnn, \neuromatch, \gen, \gmn~and \isoedge, we use the margins determined by \citet{RoyVCD2022IsoNet} for each dataset. For \isonode, the margins prescribed for \isoedge~were used for standardization. For \ournode, \ouredge~and ablations, a fixed margin of $0.5$ is used.

Procedure for baselines \textbf{\egsc, \greed, \hmn}: They are trained on five seeds with a margin of 0.5 for $10$ epochs and the best seed is chosen using the validation MAP score at this point. This seed is also used to train a model with a margin of 0.1 for $10$ epochs. The better of these models, again using MAP score on the validation split, is identified and retrained till completion using early stopping.

\begin{table}[h!]
\centering
\adjustbox{max width=\hsize}{\tabcolsep 7pt 
\begin{tabular}{l|c c c c c c}
\toprule
& \aids & \mutag & \fm & \fr& \mm & \mr  \\
\midrule \midrule 
\graphsim~\cite{graphsim} & 0.5 & 0.5 & 0.5 & 0.5 & 0.5 & 0.5\\
\gotsim~\cite{gotsim} & 0.1 & 0.1 & 0.1 & 0.1 & 0.1 & 0.1\\
\simgnn~\cite{simgnn} & 0.5 & 0.1 & 0.5 & 0.1 & 0.5 & 0.5\\
\egsc~\cite{egsc} & 0.1 & 0.5 & 0.1 & 0.5 & 0.1 & 0.5\\
\hmn~\cite{zhang2021h2mn} & 0.5 & 0.5 & 0.5 & 0.5 & 0.5 & 0.1\\
\neuromatch~\cite{neuromatch} & 0.5 & 0.5 & 0.5 & 0.5 & 0.5 & 0.5\\
\greed~\cite{greed} & 0.5 & 0.5 & 0.5 & 0.5 & 0.5 & 0.5\\
\gen~\cite{gmn} & 0.5 & 0.5 & 0.5 & 0.5 & 0.5 & 0.5\\
\gmn~\cite{gmn} & 0.5 & 0.5 & 0.5 & 0.5 & 0.5 & 0.5\\
\isonode~\cite{RoyVCD2022IsoNet} & 0.5 & 0.5 & 0.5 & 0.5 & 0.5 & 0.5\\
\isoedge~\cite{RoyVCD2022IsoNet} & 0.5 & 0.5 & 0.5 & 0.5 & 0.5 & 0.5\\
\hline \hline
 \end{tabular}}
 \vspace{1mm}
\caption{Best margin for baselines used in comparison.}
\label{tab:best-margin}
\end{table}


\subsection{Software and Hardware}
\label{app:software-hardware}
All experiments were run with Python $3.10.13$ and PyTorch $2.1.2$. \ournode, \ouredge, \gmn, \isoedge~and ablations on top of these were trained on Nvidia RTX A6000 (48 GB) GPUs while other baselines like \graphsim, \gotsim~etc. were trained on Nvidia A100 (80 GB) GPUs.

As an estimate of training time, we typically spawn $3$ training runs of \ournode or \ouredge on one Nvidia RTX A6000 GPU, each of which takes 300 epochs to conclude on average, with an average of 6-12 minutes per epoch. This amounts to $2$ days of training. Overloading the GPUs by spawning $6$ training runs per GPU increases the training time marginally to $2.5$ days.

Additionally, we use wandb~\cite{wandb} to manage and monitor the experiments.

\subsection{License}
\label{app:license}
\gen, \gmn, \gotsim, \greed~and \egsc~are available under the MIT license, while \simgnn is public under the GNU license. The licenses for \graphsim, \hmn, \isonode, \isoedge, \neuromatch~could not be identified. The authors were unable to identify the license of the TUDatasets repository~\cite{Morris+2020}, which was used to compile the $6$ datasets used in this paper.

\newpage
\section{Additional experiments}
\label{app:expts}
\subsection{Comparison against baselines}
\label{subsec:comparison_baseline}
In Tables~\ref{tab:main-app-map-hits} and~\ref{tab:main-app-mrr-precision}, we report the Mean Average Precision (MAP), HITS@20, MRR and Precision@20 scores for several baselines as well as the four approaches discussed in our paper - multi-layer and multi-round variants of \ournode and \ouredge. Multi-round \ouredge outperforms all other models with respect to all metrics, closely followed by multi-round \ournode and multi-layer \ouredge respectively. Among the baselines, \isoedge~is the best-performing model, closely followed by \isonode~and \gmn.

For MRR, Precision@20, the comparisons are less indicative of the significant boost in performance obtained by \our, since these are not aggressive metrics from the point of view of information retrieval.

\begin{table}[h!]
\centering
\maxsizebox{0.99\hsize}{!}{  \tabcolsep 7pt 
\begin{tabular}{l|c c c c c c }
\toprule
\multicolumn{7}{c}{Mean Average Precision (MAP)} \\ \hline
& \aids & \mutag & \fm & \fr& \mm & \mr  \\
\midrule \midrule 
\graphsim~\cite{graphsim}  & 0.356 $\pm$ 0.016 & 0.472 $\pm$ 0.027 & 0.477 $\pm$ 0.016 & 0.423 $\pm$ 0.019 & 0.415 $\pm$ 0.017 & 0.453 $\pm$ 0.018 \\[1ex] 
\gotsim~\cite{gotsim} & 0.324 $\pm$ 0.015 & 0.272 $\pm$ 0.012 & 0.355 $\pm$ 0.014 & 0.373 $\pm$ 0.018 & 0.323 $\pm$ 0.015 & 0.317 $\pm$ 0.013\\[1ex] 
\simgnn~\cite{simgnn}  & 0.341 $\pm$ 0.019 & 0.283 $\pm$ 0.012 & 0.473 $\pm$ 0.016 & 0.341 $\pm$ 0.015 & 0.298 $\pm$ 0.012 & 0.379 $\pm$ 0.015\\[1ex] 
\egsc~\cite{egsc}  & 0.505 $\pm$ 0.02 & 0.476 $\pm$ 0.022 & 0.609 $\pm$ 0.018 & 0.607 $\pm$ 0.019 & 0.586 $\pm$ 0.019 & 0.58 $\pm$ 0.018\\[1ex] 
\hmn~\cite{zhang2021h2mn}  & 0.267 $\pm$ 0.014 & 0.276 $\pm$ 0.012 & 0.436 $\pm$ 0.015 & 0.412 $\pm$ 0.016 & 0.312 $\pm$ 0.014 & 0.243 $\pm$ 0.008\\ [1ex] \hline 
\neuromatch~\cite{neuromatch}  & 0.489 $\pm$ 0.024 & 0.576 $\pm$ 0.029 & 0.615 $\pm$ 0.019 & 0.559 $\pm$ 0.024 & 0.519 $\pm$ 0.02 & 0.606 $\pm$ 0.021\\[1ex] 
\greed~\cite{greed}  & 0.472 $\pm$ 0.021 & 0.567 $\pm$ 0.027 & 0.558 $\pm$ 0.02 & 0.512 $\pm$ 0.021 & 0.546 $\pm$ 0.021 & 0.528 $\pm$ 0.019\\[1ex] 
\gen~\cite{gmn}  & 0.557 $\pm$ 0.021 & 0.605 $\pm$ 0.028 & 0.661 $\pm$ 0.021 & 0.575 $\pm$ 0.02 & 0.539 $\pm$ 0.02 & 0.631 $\pm$ 0.018\\[1ex] 
\gmn~\cite{gmn}  & 0.622 $\pm$ 0.02 & \lfirst{0.710 $\pm$ 0.025} & 0.730 $\pm$ 0.018 & 0.662 $\pm$ 0.02 & 0.655 $\pm$ 0.02 & 0.708 $\pm$ 0.017\\[1ex] 
\isonode~\cite{RoyVCD2022IsoNet}  & 0.659 $\pm$ 0.022 & 0.697 $\pm$ 0.026 & 0.729 $\pm$ 0.018 & 0.68 $\pm$ 0.022 & 0.708 $\pm$ 0.016 & 0.738 $\pm$ 0.017\\[1ex] 
\isoedge~\cite{RoyVCD2022IsoNet}  & \lfirst{0.690} $\pm$ 0.02 & 0.706 $\pm$ 0.026 & \lfirst{0.783 $\pm$ 0.017} & \lfirst{0.722 $\pm$ 0.02} & \lfirst{0.753 $\pm$ 0.015} & \lfirst{0.774 $\pm$ 0.016}\\[1ex] 
\hline
 \ournodelayershort & 0.756 $\pm$ 0.019 & 0.81 $\pm$ 0.021 & 0.859 $\pm$ 0.015 & 0.802 $\pm$ 0.018 & 0.827 $\pm$ 0.015 & 0.841 $\pm$ 0.013 \\[1ex] 
\ouredgelayershort & 0.795 $\pm$ 0.018 & 0.805 $\pm$ 0.022 & 0.883 $\pm$ 0.013 & 0.812 $\pm$ 0.016 & \second{0.862 $\pm$ 0.013} & \second{0.886 $\pm$ 0.011}  \\[1ex]  \hline
\ournoderoundshort & \second{0.825 $\pm$ 0.016} & \second{0.851 $\pm$ 0.018} & \second{0.888 $\pm$ 0.012} & \second{0.855 $\pm$ 0.015} & 0.838 $\pm$ 0.015 & 0.874 $\pm$ 0.011\\[1ex] 
\ouredgeroundshort & \first{0.847 $\pm$ 0.016} & \first{0.858 $\pm$ 0.019} & \first{0.902 $\pm$ 0.012} & \first{0.875 $\pm$ 0.014} & \first{0.902 $\pm$ 0.01} & \first{0.902 $\pm$ 0.01}\\ [1ex] \hline\hline
 \end{tabular}
 }
 
 \bigskip

\maxsizebox{0.99\hsize}{!}{  \tabcolsep 7pt 
\begin{tabular}{l|c c c c c c }
\toprule
\multicolumn{7}{c}{HITS@20} \\ \hline
& \aids & \mutag & \fm & \fr& \mm & \mr  \\
\midrule \midrule 
\graphsim~\cite{graphsim} & 0.145 $\pm$ 0.011 & 0.257 $\pm$ 0.027 & 0.261 $\pm$ 0.015 & 0.227 $\pm$ 0.015 & 0.212 $\pm$ 0.014 & 0.23 $\pm$ 0.015\\[1ex] 
\gotsim~\cite{gotsim} & 0.112 $\pm$ 0.011 & 0.088 $\pm$ 0.009 & 0.147 $\pm$ 0.011 & 0.166 $\pm$ 0.014 & 0.119 $\pm$ 0.011 & 0.116 $\pm$ 0.011\\[1ex] 
\simgnn~\cite{simgnn} & 0.138 $\pm$ 0.016 & 0.087 $\pm$ 0.008 & 0.235 $\pm$ 0.015 & 0.155 $\pm$ 0.013 & 0.111 $\pm$ 0.009 & 0.160 $\pm$ 0.013\\[1ex] 
\egsc~\cite{egsc} & 0.267 $\pm $0.023 & 0.243 $\pm $0.02 & 0.364 $\pm $0.02 & 0.382 $\pm $0.024 & 0.348 $\pm $0.023 & 0.325 $\pm $0.021\\[1ex] 
\hmn~\cite{zhang2021h2mn} & 0.076 $\pm$ 0.009 & 0.084 $\pm$ 0.007 & 0.200 $\pm$ 0.012 & 0.189 $\pm$ 0.013 & 0.119 $\pm$ 0.011 & 0.069
$\pm$ 0.004\\ [1ex] \hline 
\neuromatch~\cite{neuromatch} & 0.262 $\pm$ 0.025 & 0.376 $\pm$ 0.034 & 0.389 $\pm$ 0.022 & 0.350 $\pm$ 0.025 & 0.282 $\pm$ 0.019 & 0.385 $\pm$ 0.025\\[1ex] 
\greed~\cite{greed} & 0.245 $\pm$ 0.025 & 0.371 $\pm$ 0.034 & 0.316 $\pm$ 0.027 & 0.287 $\pm$ 0.019 & 0.311 $\pm$ 0.024 & 0.277 $\pm$ 0.023\\[1ex] 
\gen~\cite{gmn} & 0.321 $\pm$ 0.026 & 0.429 $\pm$ 0.035 & 0.448 $\pm$ 0.03 & 0.368 $\pm$ 0.026 & 0.292 $\pm$ 0.024 & 0.391 $\pm$ 0.025\\[1ex] 
\gmn~\cite{gmn} & 0.397 $\pm$ 0.029 & \lfirst{0.544 $\pm$ 0.035} & 0.537 $\pm$ 0.027 & 0.45 $\pm$ 0.027 & 0.423 $\pm$ 0.025 & 0.49 $\pm$ 0.026\\[1ex] 
\isonode~\cite{RoyVCD2022IsoNet} & 0.438 $\pm$ 0.028 & 0.509 $\pm$ 0.034 & 0.525 $\pm$ 0.026 & 0.475 $\pm$ 0.03 & 0.493 $\pm$ 0.023 & 0.532 $\pm$ 0.025\\[1ex] 
\isoedge~\cite{RoyVCD2022IsoNet} & \lfirst{0.479 $\pm$ 0.029} & 0.529 $\pm$ 0.035 & \lfirst{0.613 $\pm$ 0.026} & \lfirst{0.538 $\pm$ 0.029} & \lfirst{0.571 $\pm$ 0.023} & \lfirst{0.601 $\pm$ 0.027}\\ [1ex] 
\hline
\ournodelayershort & 0.57 $\pm$ 0.029 & 0.672 $\pm$ 0.033 & 0.744 $\pm$ 0.027 & 0.657 $\pm$ 0.031 & 0.68 $\pm$ 0.025 & 0.707 $\pm$ 0.024 \\[1ex] 
\ouredgelayershort & 0.626 $\pm$ 0.029 & 0.671 $\pm$ 0.035 & 0.775 $\pm$ 0.026 & 0.67 $\pm$ 0.028 & \second{0.743 $\pm$ 0.024} & \second{0.776 $\pm$ 0.021} \\ [1ex] \hline
\ournoderoundshort & \second{0.672 $\pm$ 0.027} & \second{0.732 $\pm$ 0.03} & \second{0.797 $\pm$ 0.024} & \second{0.737 $\pm$ 0.026} & 0.702 $\pm$ 0.025 & 0.755 $\pm$ 0.022\\[1ex] 
\ouredgeroundshort & \first{0.705 $\pm$ 0.028} & \first{0.749 $\pm$ 0.032} & \first{0.813 $\pm$ 0.023} & \first{0.769 $\pm$ 0.026} & \first{0.809 $\pm$ 0.019} & \first{0.803 $\pm$ 0.02}\\[1ex]
 \hline \hline  
 \end{tabular}
 }
 \vspace{1mm}
\caption{Replication of Table~\ref{tab:main} with standard error. Comparison of the two variants of \our\ (\ournode and \ouredge) against all the state-of-the-art graph retrieval methods, across all six datasets. Performance is measured in terms average precision MAP and HITS@20. In all cases, we used 60\% training, 15\% validation and 25\% test sets. The first five methods apply a neural network on the fused graph-pair representations. The next six methods apply asymmetric hinge distance between the query and corpus embeddings similar to our method. The numbers with \fboxg{green} and \fboxy{yellow} indicate the best, second best method respectively, whereas the numbers with \fboxb{blue} indicate the best method among the baselines.  (MAP values for \ouredge across \fm, \mm\ and \mr\ are verified to be not exactly same, but they take the same value until the third decimal).}
\label{tab:main-app-map-hits}
\end{table}

\begin{table}[h!]
\centering
\maxsizebox{0.99\hsize}{!}{  \tabcolsep 7pt 
\begin{tabular}{l|c c c c c c }
\toprule
\multicolumn{7}{c}{Mean Reciprocal Rank (MRR)} \\ \hline 
& \aids & \mutag & \fm & \fr& \mm & \mr  \\
\midrule \midrule 
\graphsim~\cite{graphsim}  & 0.71 $\pm $0.039 & 0.795 $\pm $0.037 & 0.885 $\pm $0.029 & 0.817 $\pm $0.032 & 0.818 $\pm $0.034 & 0.789 $\pm $0.037 \\[1ex] 
\gotsim~\cite{gotsim} & 0.568 $\pm $0.038 & 0.584 $\pm $0.037 & 0.775 $\pm $0.037 & 0.716 $\pm $0.042 & 0.459 $\pm $0.045 & 0.525 $\pm $0.047\\[1ex] 
\simgnn~\cite{simgnn}  & 0.533 $\pm $0.038 & 0.644 $\pm $0.043 & 0.866 $\pm $0.031 & 0.753 $\pm $0.038 & 0.669 $\pm $0.04 & 0.638 $\pm $0.046\\[1ex] 
\egsc~\cite{egsc}  & 0.894 $\pm $0.026 & 0.75 $\pm $0.041 & 0.943 $\pm $0.021 & 0.909 $\pm $0.023 & 0.904 $\pm $0.025 & 0.932 $\pm $0.022\\[1ex] 
\hmn~\cite{zhang2021h2mn} & 0.46 $\pm $0.047 & 0.565 $\pm $0.042 & 0.822 $\pm $0.035 & 0.817 $\pm $0.034 & 0.386 $\pm $0.039 & 0.62 $\pm $0.041\\ [1ex] \hline 
\neuromatch~\cite{neuromatch}  & 0.823 $\pm $0.035 & 0.855 $\pm $0.035 & 0.88 $\pm $0.028 & 0.929 $\pm $0.022 & 0.87 $\pm $0.027 & 0.895 $\pm $0.026\\[1ex] 
\greed~\cite{greed}  & 0.789 $\pm $0.035 & 0.805 $\pm $0.034 & 0.834 $\pm $0.033 & 0.834 $\pm $0.032 & 0.894 $\pm $0.028 & 0.759 $\pm $0.039\\[1ex] 
\gen~\cite{gmn}  & 0.865 $\pm $0.028 & 0.895 $\pm $0.029 & 0.889 $\pm $0.026 & 0.878 $\pm $0.028 & 0.814 $\pm $0.034 & 0.878 $\pm $0.026\\[1ex] 
\gmn~\cite{gmn}  & 0.877 $\pm $0.027 & 0.923 $\pm $0.023 & 0.949 $\pm $0.019 & 0.947 $\pm $0.019 & 0.928 $\pm $0.023 & 0.922 $\pm $0.022\\[1ex] 
\isonode~\cite{RoyVCD2022IsoNet}  & 0.916 $\pm $0.024 & 0.887 $\pm $0.029 & \lfirst{0.977 $\pm $0.013} & 0.954 $\pm $0.018 & 0.956 $\pm $0.018 & \lfirst{0.954 $\pm $0.018}\\[1ex] 
\isoedge~\cite{RoyVCD2022IsoNet}  & \lfirst{0.949 $\pm $0.02} & \lfirst{0.926 $\pm $0.026} & 0.973 $\pm $0.013 & \lfirst{0.956 $\pm $0.018} & \lfirst{0.98 $\pm $0.011} & 0.948 $\pm $0.019\\[1ex] 
\hline
 \ournodelayershort & 0.956 $\pm $0.018 & 0.954 $\pm $0.018 & \first{1.0 $\pm $0.0} & 0.978 $\pm $0.013 & 0.98 $\pm $0.011 & \first{1.0 $\pm $0.0} \\[1ex] 
\ouredgelayershort & 0.984 $\pm $0.011 & \second{0.976 $\pm $0.014} & \second{0.991 $\pm $0.009} & 0.987 $\pm $0.009 & 0.987 $\pm $0.009 & \second{0.993 $\pm $0.007}\\[1ex]  \hline
\ournoderoundshort & \second{0.993 $\pm $0.007} & 0.971 $\pm $0.014 & \first{1.0 $\pm $0.0} & \second{0.993 $\pm $0.007} & \second{0.993 $\pm $0.007} & \second{0.993 $\pm $0.007}\\[1ex] 
\ouredgeroundshort & \first{1.0 $\pm $0.0} & \first{0.983 $\pm $0.012} & \second{0.991 $\pm $0.009} & \first{1.0 $\pm $0.0} & \first{1.0 $\pm $0.0} & \first{1.0 $\pm $0.0} \\ [1ex] \hline\hline
 \end{tabular}
 }
 
 \bigskip

\maxsizebox{0.99\hsize}{!}{  \tabcolsep 7pt 
\begin{tabular}{l|c c c c c c }
\toprule
\multicolumn{7}{c}{Precision@20} \\ \hline
& \aids & \mutag & \fm & \fr& \mm & \mr  \\
\midrule \midrule 
\graphsim~\cite{graphsim} & 0.474 $\pm $0.025 & 0.577 $\pm $0.033 & 0.679 $\pm $0.023 & 0.617 $\pm $0.028 & 0.604 $\pm $0.028 & 0.638 $\pm $0.026\\[1ex] 
\gotsim~\cite{gotsim} & 0.386 $\pm $0.024 & 0.325 $\pm $0.021 & 0.479 $\pm $0.027 & 0.519 $\pm $0.03 & 0.409 $\pm $0.027 & 0.421 $\pm $0.03\\[1ex] 
\simgnn~\cite{simgnn} & 0.44 $\pm $0.026 & 0.33 $\pm $0.022 & 0.626 $\pm $0.026 & 0.471 $\pm $0.029 & 0.414 $\pm $0.026 & 0.512 $\pm $0.032\\[1ex] 
\egsc~\cite{egsc} & 0.646 $\pm $0.023 & 0.608 $\pm $0.034 & 0.79 $\pm $0.022 & 0.766 $\pm $0.021 & 0.739 $\pm $0.023 & 0.74 $\pm $0.021\\[1ex] 
\hmn~\cite{zhang2021h2mn} & 0.28 $\pm $0.026 & 0.34 $\pm $0.023 & 0.587 $\pm $0.024 & 0.563 $\pm $0.026 & 0.399 $\pm $0.028 & 0.308 $\pm $0.017\\ [1ex] \hline 
\neuromatch~\cite{neuromatch} & 0.615 $\pm $0.03 & 0.689 $\pm $0.032 & 0.809 $\pm $0.022 & 0.725 $\pm $0.027 & 0.694 $\pm $0.027 & 0.751 $\pm $0.023\\[1ex] 
\greed~\cite{greed} & 0.591 $\pm $0.024 & 0.661 $\pm $0.03 & 0.689 $\pm $0.026 & 0.642 $\pm $0.028 & 0.699 $\pm $0.028 & 0.624 $\pm $0.029\\[1ex] 
\gen~\cite{gmn} & 0.674 $\pm $0.024 & 0.721 $\pm $0.03 & 0.783 $\pm $0.023 & 0.678 $\pm $0.022 & 0.64 $\pm $0.027 & 0.759 $\pm $0.021\\[1ex] 
\gmn~\cite{gmn} & 0.751 $\pm $0.022 & \lfirst{0.82 $\pm $0.023} & 0.852 $\pm $0.02 & 0.809 $\pm $0.019 & 0.783 $\pm $0.022 & 0.832 $\pm $0.018\\[1ex] 
\isonode~\cite{RoyVCD2022IsoNet} & 0.791 $\pm $0.022 & 0.803 $\pm $0.029 & 0.866 $\pm $0.018 & 0.803 $\pm $0.022 & 0.844 $\pm $0.015 & 0.863 $\pm $0.016\\[1ex] 
\isoedge~\cite{RoyVCD2022IsoNet} & \lfirst{0.822 $\pm $0.022} & 0.812 $\pm $0.028 & \lfirst{0.896 $\pm $0.016} & \lfirst{0.851 $\pm $0.017} & \lfirst{0.877 $\pm $0.014} & \lfirst{0.875 $\pm $0.017}\\ [1ex] 
\hline
\ournodelayershort & 0.873 $\pm $0.018 & 0.897 $\pm $0.018 & 0.935 $\pm $0.012 & 0.917 $\pm $0.012 & 0.93 $\pm $0.013 & 0.931 $\pm $0.012 \\[1ex] 
\ouredgelayershort & 0.905 $\pm $0.015 & 0.883 $\pm $0.021 & \second{0.958 $\pm $0.01} & 0.93 $\pm $0.01 & \second{0.953 $\pm $0.01} & \second{0.976 $\pm $0.005}\\ [1ex] \hline
\ournoderoundshort & \second{0.932 $\pm $0.012} & \first{0.943 $\pm $0.011} & 0.957 $\pm $0.01 & \second{0.961 $\pm $0.008} & 0.949 $\pm $0.011 & 0.963 $\pm $0.008\\[1ex] 
\ouredgeroundshort & \first{0.946 $\pm $0.012} & \second{0.931 $\pm $0.014} & \first{0.973 $\pm $0.007} & \first{0.963 $\pm $0.008} & \first{0.98 $\pm $0.005} & \first{0.987 $\pm $0.003}\\[1ex]
 \hline \hline  
 \end{tabular}
 }
 \vspace{1mm}
\caption{MRR and Precision@20 of corresponding models from Table~\ref{tab:main} with standard error. Comparison of the two variants of \our\ (\ournode and \ouredge) against all the state-of-the-art graph retrieval methods, across all six datasets. Performance is measured in terms MRR and Precision@20. In all cases, we used 60\% training, 15\% validation and 25\% test sets. The first five methods apply a neural network on the fused graph-pair representations. The next six methods apply asymmetric hinge distance between the query and corpus embeddings similar to our method. The numbers with \fboxg{green} and \fboxy{yellow} indicate the best, second best method respectively, whereas the numbers with \fboxb{blue} indicate the best method among the baselines.}
\label{tab:main-app-mrr-precision}
\end{table} 

\subsection{HITS@20, MRR and Precision@20 for multi-round \our\ and multi-layer \our}
Table~\ref{tab:multi-app} compares multi-round and multi-layer \our with respect to different metrics. We observe that multi-round \our outperforms multi-layer \our by a significant margin when it comes to all metrics, both when the models are node-based or edge-based. This reinforces the observations from MAP scores noted earlier in Table~\ref{tab:multi}.
Note that a minor exception occurs for MRR but the scores are already so close to $1$ that this particular metric can be discounted and our key observation above still stands.

\begin{table}[!!!h]
\centering
\maxsizebox{0.9\hsize}{!}{
\small
\begin{tabular}{l|ccc ccc}
\hline
\multicolumn{7}{c}{HITS@20} \\ \hline

& \aids & \mutag & \fm & \fr & \mm & \mr\\ \hline\hline
\multirow{2}{0.1cm}{\hspace{-0.5cm}\begin{sideways}{\textbf{Node}}\end{sideways}}\hspace{-4mm}\ldelim\{{2}{0.3mm}\hspace{2mm}
  Multi-layer & 0.57 & 0.672 & 0.744 & 0.657 & 0.68 & 0.707 \\[1ex]
 Multi-round & \second{0.672} & \second{0.732} & \second{0.797} & \second{0.737} & 0.702 & 0.755 \\\hline\hline
  \multirow{2}{0.1cm}{\hspace{-0.5cm}\begin{sideways}{\textbf{Edge}}\end{sideways}}\hspace{-4mm}\ldelim\{{2}{0.3mm}\hspace{2mm}
   Multi-layer  & 0.626 & 0.671 & 0.775 & 0.67 & \second{0.743} & \second{0.776} \\[1ex] 
 Multi-round  & \first{0.705} & \first{0.749} & \first{0.813} & \first{0.769} & \first{0.809} & \first{0.803} \\ \hline\hline
 \end{tabular}}

 \bigskip

 \maxsizebox{0.9\hsize}{!}{
\small
\begin{tabular}{l|ccc ccc}
\hline
\multicolumn{7}{c}{Mean Reciprocal Rank (MRR)} \\ \hline 
& \aids & \mutag & \fm & \fr & \mm & \mr\\ \hline\hline
\multirow{2}{0.1cm}{\hspace{-0.5cm}\begin{sideways}{\textbf{Node}}\end{sideways}}\hspace{-4mm}\ldelim\{{2}{0.3mm}\hspace{2mm}
  Multi-layer & 0.956 & 0.954 & \first{1.0}  & 0.978  & 0.98  & \first{1.0} \\[1ex]
 Multi-round & 0.993 & 0.971 & \first{1.0} & \second{0.993} & \second{0.993} & \second{0.993} \\\hline\hline
  \multirow{2}{0.1cm}{\hspace{-0.5cm}\begin{sideways}{\textbf{Edge}}\end{sideways}}\hspace{-4mm}\ldelim\{{2}{0.3mm}\hspace{2mm}
   Multi-layer & 0.984 & \second{0.976} & \second{0.991} & 0.987 & 0.987  & \second{0.993} \\[1ex] 
 Multi-round  &  \first{1.0} & \first{0.983}  & \second{0.991}  & \first{1.0}  & \first{1.0} & \first{1.0}\\ \hline\hline
 \end{tabular}}

\bigskip

 \maxsizebox{0.9\hsize}{!}{
\small
\begin{tabular}{l|ccc ccc}
\hline
\multicolumn{7}{c}{Precision@20} \\ \hline

& \aids & \mutag & \fm & \fr & \mm & \mr\\ \hline\hline
\multirow{2}{0.1cm}{\hspace{-0.5cm}\begin{sideways}{\textbf{Node}}\end{sideways}}\hspace{-4mm}\ldelim\{{2}{0.3mm}\hspace{2mm}
  Multi-layer & 0.873 & 0.897 & 0.935 & 0.917 & 0.93 & 0.931 \\[1ex]
 Multi-round & \second{0.932} & \first{0.943} & 0.957 & \second{0.961} & 0.949 & 0.963 \\\hline\hline
  \multirow{2}{0.1cm}{\hspace{-0.5cm}\begin{sideways}{\textbf{Edge}}\end{sideways}}\hspace{-4mm}\ldelim\{{2}{0.3mm}\hspace{2mm}
   Multi-layer & 0.905  & 0.883 & \second{0.958}  & 0.93  & \second{0.953}  & \second{0.976} \\[1ex] 
 Multi-round  &  \first{0.946} & \second{0.931} & \first{0.973}  & \first{0.963} & \first{0.98}  & \first{0.987}  \\ \hline\hline
 \end{tabular}}
 \vspace{1mm}
\caption{\small Multi-round vs. multi-layer refinement. First and the last two rows of each table report HITS@20, MRR and Precision@20 for \ournode\ and \ouredge respectively. Rows colored \fboxg{green} and \fboxy{yellow} indicate the best and second best methods respectively.}
\label{tab:multi-app}
\end{table}


\subsection{Refinement of alignment matrix across rounds and layers in \ourround\ and  multi-layer \our}
\label{subsec:round-histograms}
\begin{figure}[h]
    \centering
    \vspace{-4mm}

\subfloat{\includegraphics[width=.24\linewidth]{FIG/NEW_HIST/aids_node_histogram.pdf}}   
\subfloat{\includegraphics[width=.24\linewidth]{FIG/NEW_HIST/aids_node_baseline_histogram.pdf}}   \hspace{3mm}
\subfloat{\includegraphics[width=.24\linewidth]{FIG/NEW_HIST/aids_edge_histogram.pdf}} 
\subfloat{\includegraphics[width=.24\linewidth]{FIG/NEW_HIST/aids_edge_baseline_histogram.pdf}}

\subfloat{\includegraphics[width=.24\linewidth]{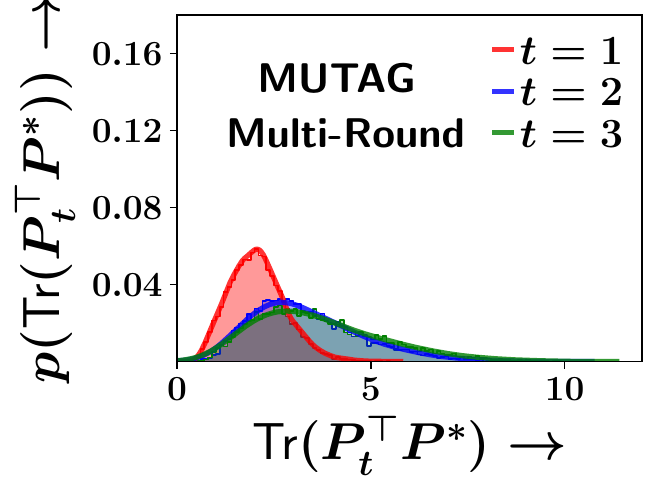}} 
\subfloat{\includegraphics[width=.24\linewidth]{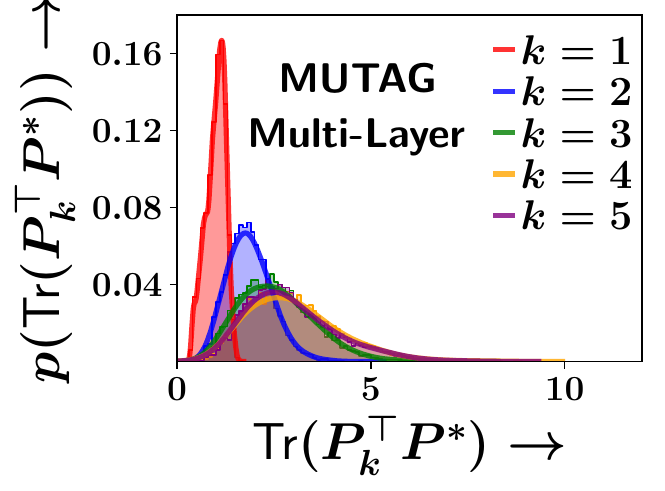}}   \hspace{3mm}
\subfloat{\includegraphics[width=.24\linewidth]{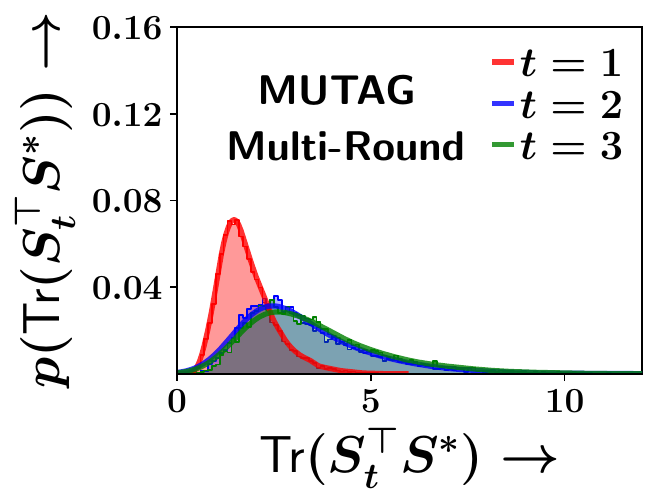}}
\subfloat{\includegraphics[width=.24\linewidth]{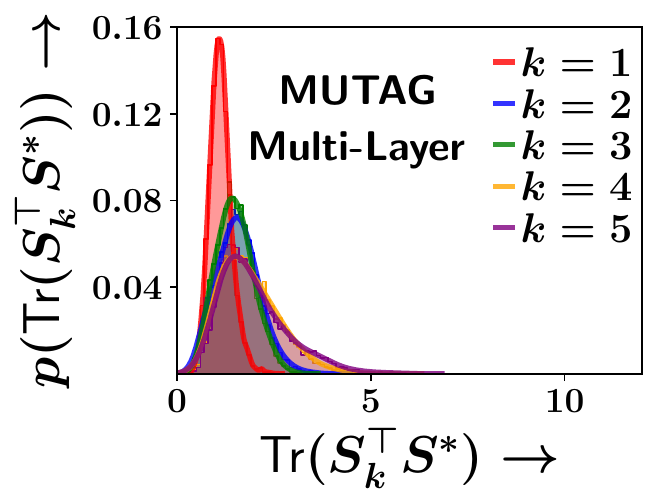}}

\subfloat{\includegraphics[width=.24\linewidth]{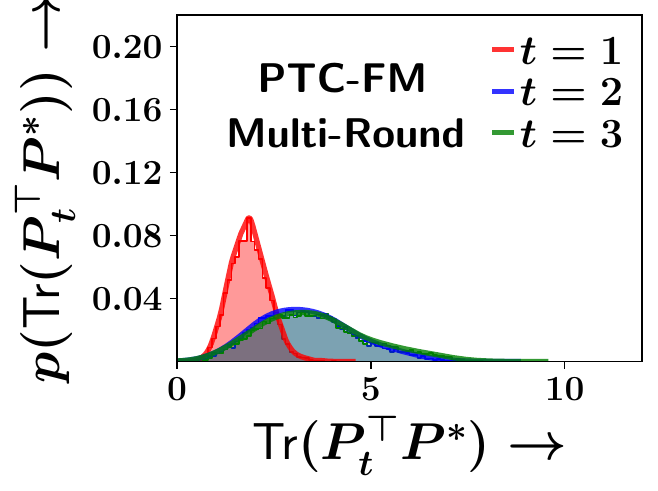}} 
\subfloat{\includegraphics[width=.24\linewidth]{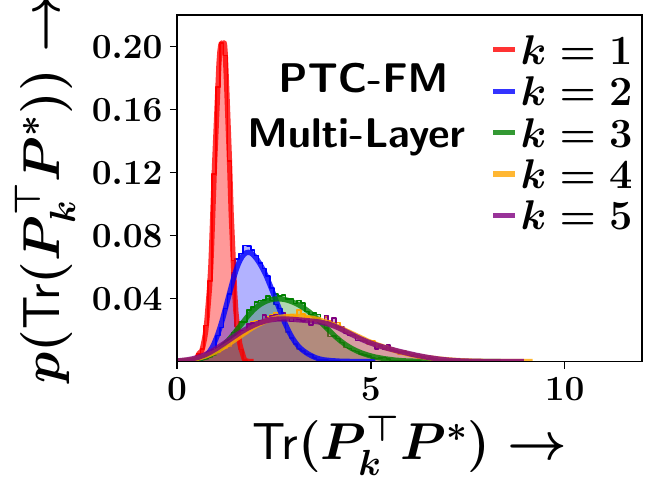}}   \hspace{3mm}
\subfloat{\includegraphics[width=.24\linewidth]{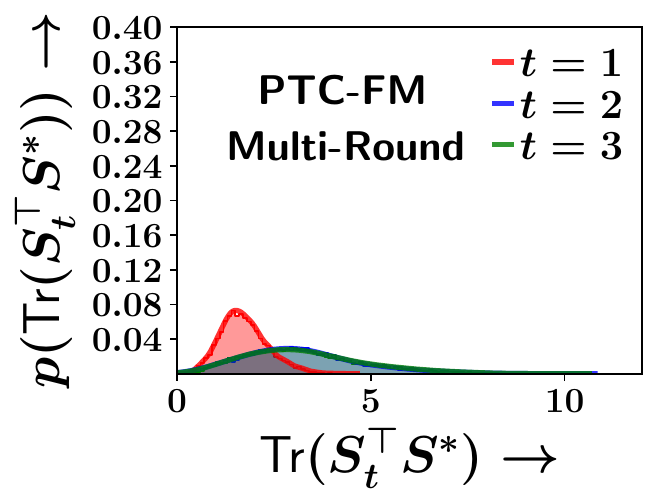}}
\subfloat{\includegraphics[width=.24\linewidth]{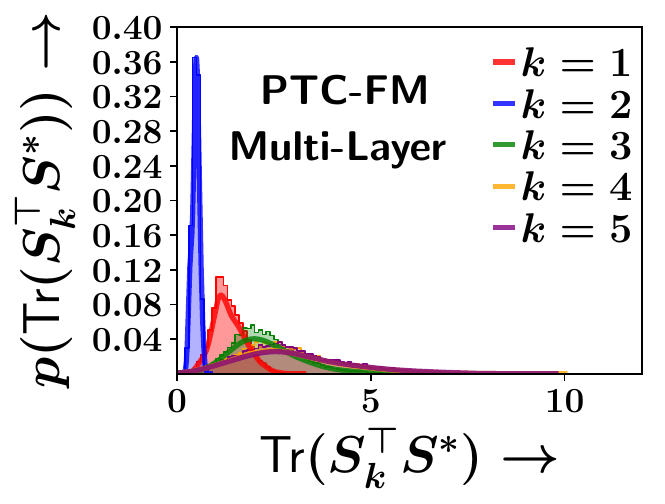}}   

\subfloat{\includegraphics[width=.24\linewidth]{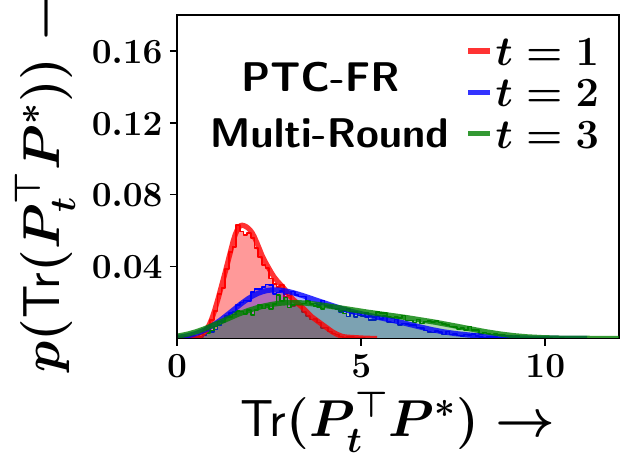}} 
\subfloat{\includegraphics[width=.24\linewidth]{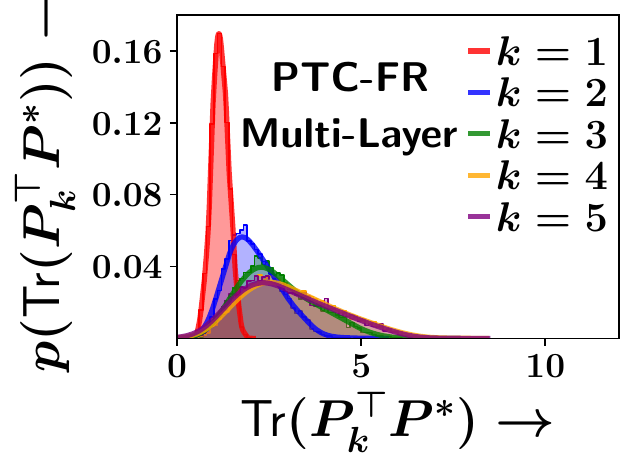}}   \hspace{3mm}
\subfloat{\includegraphics[width=.24\linewidth]{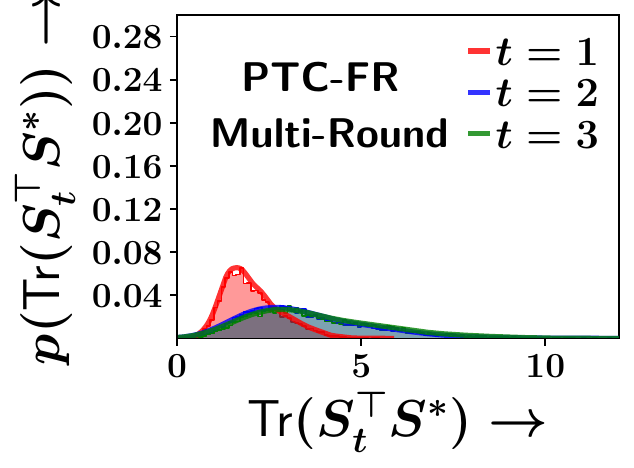}}
\subfloat{\includegraphics[width=.24\linewidth]{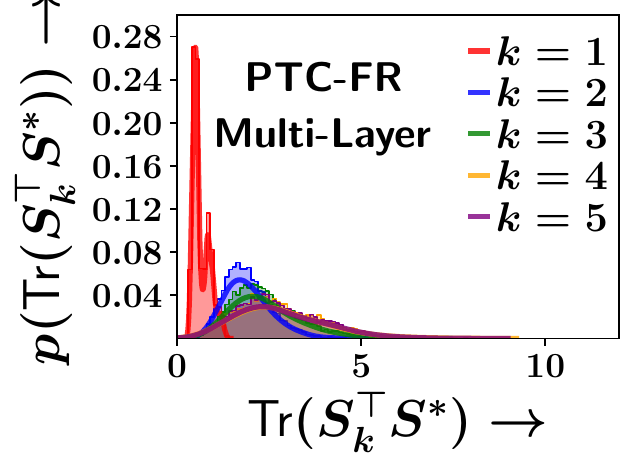}}

\subfloat{\includegraphics[width=.24\linewidth]{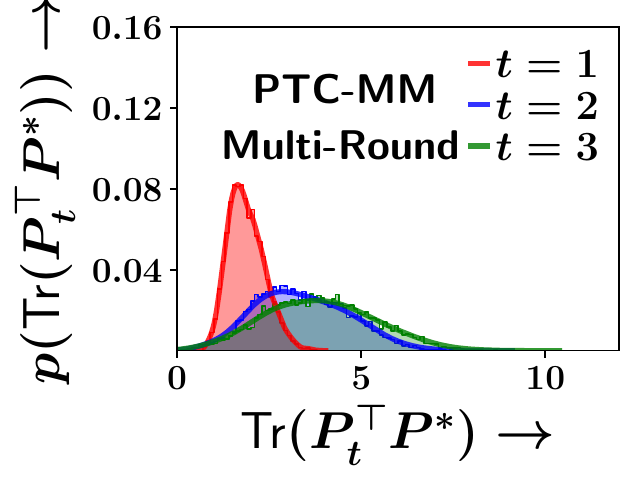}} 
\subfloat{\includegraphics[width=.24\linewidth]{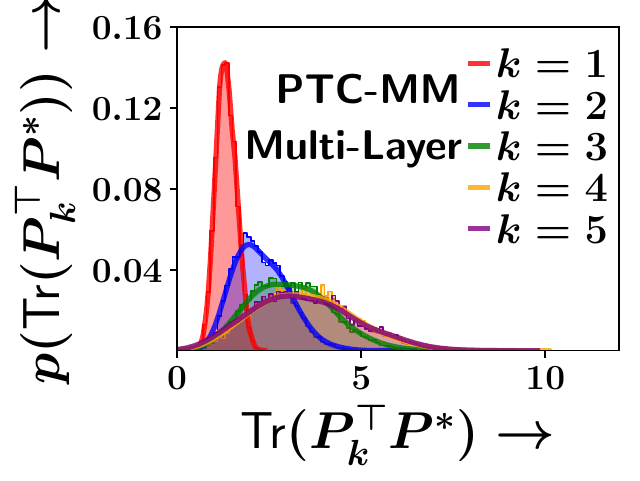}}   \hspace{3mm}
\subfloat{\includegraphics[width=.24\linewidth]{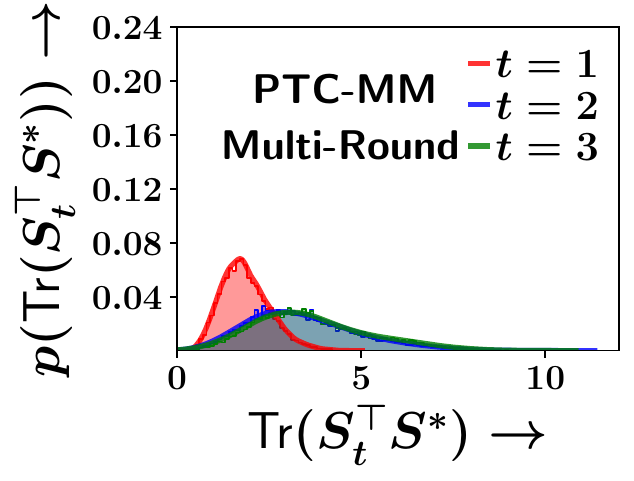}}
\subfloat{\includegraphics[width=.24\linewidth]{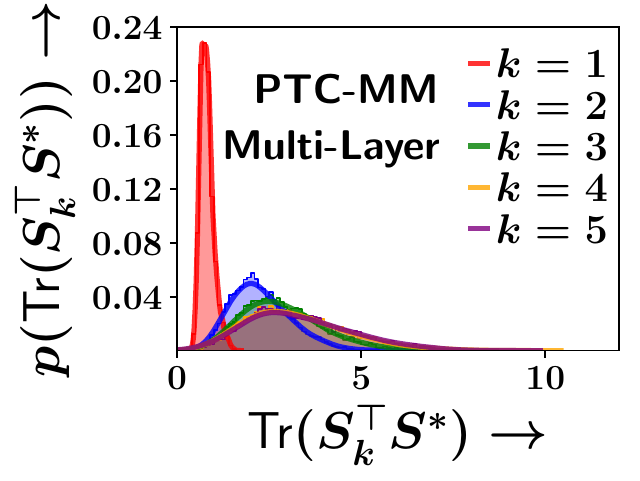}}   

\subfloat{\includegraphics[width=.24\linewidth]{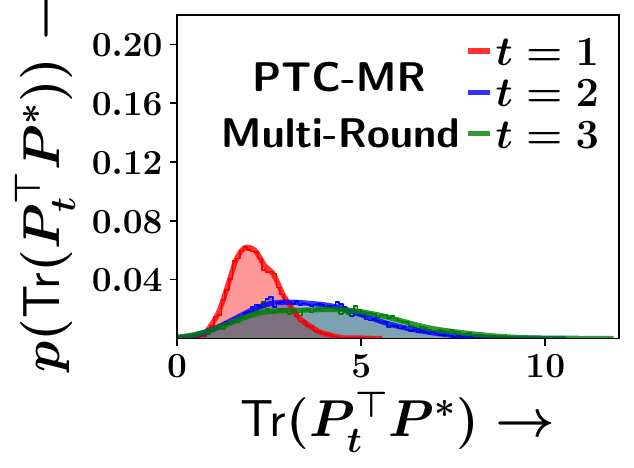}} 
\subfloat{\includegraphics[width=.24\linewidth]{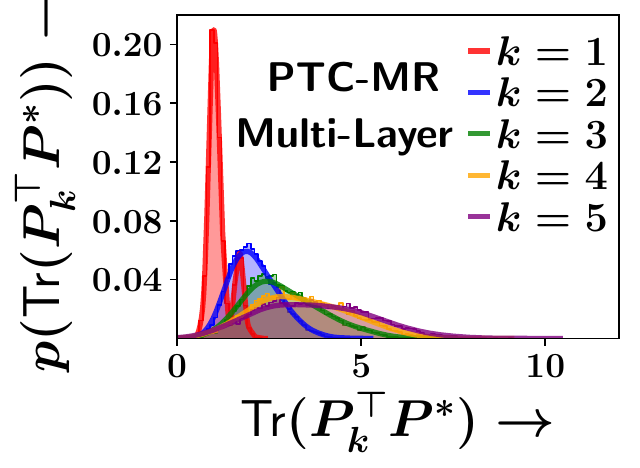}}   \hspace{3mm}
\subfloat{\includegraphics[width=.24\linewidth]{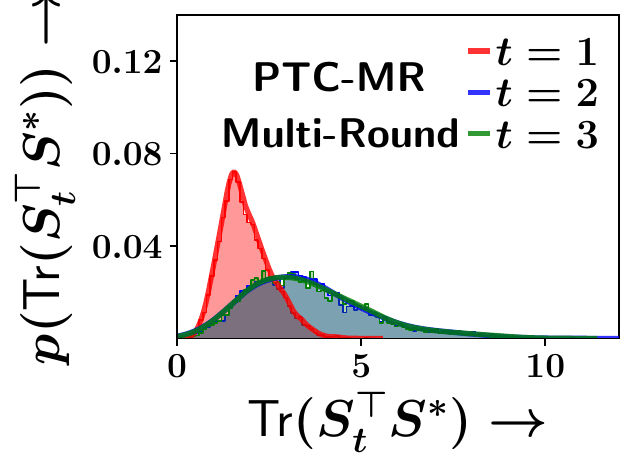}} 
\subfloat{\includegraphics[width=.24\linewidth]{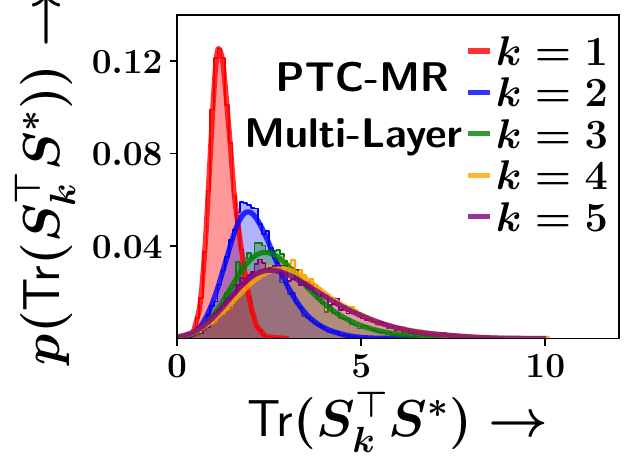}}  

\caption{Similar to Figure~\ref{fig:hist}, we plot empirical probability density of $p(\text{Tr}(\Pb_t ^{\top}\Pb^*))$ and $p(\text{Tr}(\Sb_t ^{\top}\Sb^*))$ for different values of $t$ lazy multi round updates and 
$p(\text{Tr}(\Pb_k ^{\top}\Pb^*))$ and $p(\text{Tr}(\Sb_k ^{\top}\Sb^*))$ for different values of $k$ 
 for eager multi layer updates.
The first (last) two plots in the left (right) of each row are for \ournoderound (\ouredgeround).}
\label{fig:hist-app-round}
\end{figure}

The node (edge) alignment calculated after round $t$ is denoted as $\Pb_t$ ($\Sb_t$). We accumulate such alignments across multiple rounds. This also includes $\Pb_T$ ($\Sb_T$) which is used to compute the relevance distance in Eq.~\ref{eq:node-score} (Eq.~\ref{eq:edge-score}).
We wish to compare the predicted alignments with ground truth alignments. We expect our final alignment matrix $\Pb_t$ ($\Sb_t$) to be one of them. We determine the closest ground truth matrices $\Pb^*$ and $\Sb^*$ by computing $\max_{\Pb}\operatorname{Tr}(\Pb_T ^{\top}\Pb)$ and $\max_{\Sb}\operatorname{Tr}(\Sb_T ^{\top}\Sb)$ for \ournode and \ouredge respectively.
We now use the closest ground-truth alignment $\Pb^*$, to compute $\operatorname{Tr}(\Pb_t ^{\top}\Pb^*)$ for $t \in [T]$. For each $t$, we plot a histogram with bin width 0.1 that denotes the density estimate $\operatorname{p}(\operatorname{Tr}(\Pb_t ^{\top}\Pb^*))$. The same procedure is adopted for edges, with $\Sb^*$ used instead of $\Pb^*$. The histograms are depicted in Figure~\ref{fig:hist-app-round}. We observe that the plots shift rightward with increasing $t$. The frequency of graph pairs with misaligned $\Pb_t$ ($\Sb_t$) decreases with rounds $t$ while that with well-aligned $\Pb_t$ ($\Sb_t$) increases.

Here, we also study  alignments obtained through multi-layer refinement. We adopt the same procedure as in Section~\ref{subsec:round-histograms}. One key difference is that the node/edge alignments are computed after every layer $k$ and are accumulated across layers $k \in [K]$.
In Figure~\ref{fig:hist-app-round}, we observe that the plots, in general, shift rightward with increasing $k$. The frequency of graph pairs with misaligned $\Pb_t$ ($\Sb_t$) decreases with rounds $k$ while that with well-aligned $\Pb_k$ ($\Sb_k$) increases.

\subsection{Comparison across alternatives of \ournodelayer and \ournoderound}
In Table~\ref{tab:cross-our}, we compare different alternatives to the multi-round and multi-layer variants of \ournode. In particular, we consider four alternatives - \post~(equation shown in Section~\ref{sec:Expt}), \uonly~[Appendix~\ref{app:uonly-equations}], \monly~[Appendix~\ref{app:monly-equations}] and \ournode. We observe that for all metrics, \ournode~and \monly~dominate the other alternatives in most cases. This highlights the importance of node pair partner interaction for determining the subgraph isomorphism relationship between two graphs. For the multi-round variant, \ournode~outperforms \monly in four of the datasets and is comparable / slightly worse in the other two. Once again, comparisons based on MRR break down because it does not cause a strong differentiation between the approaches.
\begin{table}[!!h]
\centering
\maxsizebox{0.9\hsize}{!}{
\small
\begin{tabular}{l|ccc ccc}
\hline
\multicolumn{7}{c}{Mean Average Precision (MAP)} \\ \hline
& \aids & \mutag & \fm & \fr & \mm & \mr\\ \hline\hline
\multirow{4}{0.1cm}{\hspace{-0.5cm}\begin{sideways}{\textbf{Multi-Layer}}\end{sideways}}\hspace{-4mm}\ldelim\{{5}{0.3mm}\hspace{2mm}
\uonly & 0.692 & 0.782 & 0.822 & 0.776 & 0.777 & 0.803 \\[1ex] 
\monly & \first{0.765} & \second{0.792} & \first{0.876} & \first{0.823} & \first{0.843} & \first{0.848} \\[1ex] 
\post & 0.668 & 0.783 & 0.821 & 0.752 & 0.753 & 0.794 \\[1ex] 
\ournode & \second{0.756} & \first{0.81} & \second{0.859} & \second{0.802} & \second{0.827} & \second{0.841}   \\ \hline\hline
\multirow{4}{0.1cm}{\hspace{-0.5cm}\begin{sideways}{\textbf{Multi-Round}}\end{sideways}}\hspace{-4mm}\ldelim\{{5}{0.3mm}\hspace{2mm}
\uonly & 0.815 & \second{0.844} & 0.868 & \second{0.852} & 0.818 & 0.858 \\[1ex] 
\monly & \second{0.818} & 0.833 & \first{0.897} & 0.831 & \first{0.852} & \second{0.871} \\[1ex] 
\post & 0.776 & 0.829 & 0.851 & 0.819 & \second{0.844} & 0.84 \\[1ex] 
\ournode & \first{0.825} & \first{0.851} & \second{0.888} & \first{0.855} & 0.838 & \first{0.874}  \\ \hline\hline
 \end{tabular}}

\medskip

\maxsizebox{0.9\hsize}{!}{
\small
\begin{tabular}{l|ccc ccc}
\hline
\multicolumn{7}{c}{HITS@20} \\ \hline
& \aids & \mutag & \fm & \fr & \mm & \mr\\ \hline\hline
\multirow{4}{0.1cm}{\hspace{-0.5cm}\begin{sideways}{\textbf{Multi-Layer}}\end{sideways}}\hspace{-4mm}\ldelim\{{5}{0.3mm}\hspace{2mm}
\uonly & 0.479 & 0.634 & 0.677 & 0.611 & 0.608 & 0.64 \\[1ex] 
\monly & \first{0.577} & \second{0.651} & \first{0.775} & \first{0.682} & \first{0.719} & \second{0.703} \\[1ex] 
\post & 0.433 & 0.639 & 0.678 & 0.58 & 0.571 & 0.624 \\[1ex] 
\ournode & \second{0.57} & \first{0.672} & \second{0.744} & \second{0.657} & \second{0.68} & \first{0.707}   \\ \hline\hline
\multirow{4}{0.1cm}{\hspace{-0.5cm}\begin{sideways}{\textbf{Multi-Round}}\end{sideways}}\hspace{-4mm}\ldelim\{{5}{0.3mm}\hspace{2mm}
\uonly & 0.658 & \second{0.727} & 0.756 & \first{0.738} & 0.667 & 0.743 \\[1ex] 
\monly & \second{0.671} & 0.717 & \first{0.807} & 0.696 & \first{0.728} & \second{0.753} \\[1ex] 
\post & 0.603 & 0.702 & 0.736 & 0.686 & \second{0.721} & 0.695 \\[1ex] 
\ournode & \first{0.672} & \first{0.732} & \second{0.797} & \second{0.737} & 0.702 & \first{0.755} \\ \hline\hline
\end{tabular}}

\medskip

\maxsizebox{0.9\hsize}{!}{
\small
\begin{tabular}{l|ccc ccc}
\hline
\multicolumn{7}{c}{Mean Reciprocal Rank (MRR)} \\ \hline 
& \aids & \mutag & \fm & \fr & \mm & \mr\\ \hline\hline
\multirow{4}{0.1cm}{\hspace{-0.5cm}\begin{sideways}{\textbf{Multi-Layer}}\end{sideways}}\hspace{-4mm}\ldelim\{{5}{0.3mm}\hspace{2mm}
\uonly & 0.909 & 0.941 & \second{0.965} & 0.964 & 0.966 & \second{0.984} \\[1ex] 
\monly & \first{0.97} & \first{0.956} & 0.964 & \first{0.993} & \second{0.978} & \first{1.0} \\[1ex] 
\post & 0.917 & 0.945 & 0.964 & \second{0.987} & 0.958 & 0.969 \\[1ex] 
\ournode & \second{0.956} & \second{0.954} & \first{1.0} & 0.978 & \first{0.98} & \first{1.0}   \\ \hline\hline
\multirow{4}{0.1cm}{\hspace{-0.5cm}\begin{sideways}{\textbf{Multi-Round}}\end{sideways}}\hspace{-4mm}\ldelim\{{5}{0.3mm}\hspace{2mm}
\uonly & \second{0.987} & 0.944 & \second{0.993} & \second{0.987} & 0.963 & 0.983 \\[1ex] 
\monly & 0.984 & \second{0.958} & \second{0.993} & 0.98 & \second{0.984} & \second{0.984} \\[1ex] 
\post & 0.984 & 0.949 & \second{0.993} & 0.978 & 0.978 & 0.97 \\[1ex] 
\ournode & \first{0.993} & \first{0.971} & \first{1.0} & \first{0.993} & \first{0.993} & \first{0.993} \\ \hline\hline
\end{tabular}}

\medskip

\maxsizebox{0.9\hsize}{!}{
\small
\begin{tabular}{l|ccc ccc}
\hline
\multicolumn{7}{c}{Precision@20} \\ \hline
& \aids & \mutag & \fm & \fr & \mm & \mr\\ \hline\hline
\multirow{4}{0.1cm}{\hspace{-0.5cm}\begin{sideways}{\textbf{Multi-Layer}}\end{sideways}}\hspace{-4mm}\ldelim\{{5}{0.3mm}\hspace{2mm}
\uonly & 0.817 & 0.867 & 0.913 & 0.913 & 0.883 & 0.914 \\[1ex] 
\monly & \second{0.871} & \second{0.886} & \first{0.957} & \first{0.937} & \second{0.927} & \first{0.937} \\[1ex] 
\post & 0.799 & 0.866 & 0.919 & 0.877 & 0.873 & 0.885 \\[1ex] 
\ournode & \first{0.873} & \first{0.897} & \second{0.935} & \second{0.917} & \first{0.93} & \second{0.931}   \\ \hline\hline
\multirow{4}{0.1cm}{\hspace{-0.5cm}\begin{sideways}{\textbf{Multi-Round}}\end{sideways}}\hspace{-4mm}\ldelim\{{5}{0.3mm}\hspace{2mm}
\uonly & 0.921 & 0.917 & 0.936 & \second{0.951} & 0.921 & 0.945 \\[1ex] 
\monly & \second{0.923} & 0.913 & \first{0.969} & \second{0.951} & \first{0.957} & \second{0.957} \\[1ex] 
\post & 0.875 & \second{0.921} & 0.933 & 0.942& 0.939 & 0.941 \\[1ex] 
\ournode & \first{0.932} & \first{0.943} & \second{0.957} & \first{0.961} & \second{0.949} & \first{0.963} \\ \hline\hline
\end{tabular}}
\caption{\small Effect of node pair partner interaction in \ournode. Table shows the comparison of \ournode with three different alternatives. The first table reports MAP values, second reports HITS@20, third reports MRR and fourth reports Precision@20. In each table, the first two rows report metrics for multi-layer refinement and the second two rows report metrics for multi-round refinement. Rows colored \fboxg{green} and \fboxy{yellow} indicate the best and second best methods in their respective sections.}
\label{tab:cross-our}
\end{table}

\clearpage
\newpage

\subsection{Comparison of \gmn with \our alternative for multi-layer and multi-round}
\label{app:monly-vs-uonly}
In Table~\ref{tab:cross-our-gmn}, we modify the GMN architecture to include node pair partner interaction in the message-passing layer. Based on the reported metrics, we observe that there is no substantial improvement upon including information from node pairs in GMN, which is driven by a non-injective mapping (attention). This indicates that injectivity of the doubly stochastic matrix in our formulation is crucial towards the boost in performance obtained from node pair partner interaction as well.

\begin{table}[!!h]
\centering
\maxsizebox{0.9\hsize}{!}{
\small
\begin{tabular}{l|ccc ccc}
\hline
\multicolumn{7}{c}{Mean Average Precision (MAP)} \\ \hline
& \aids & \mutag & \fm & \fr & \mm & \mr\\ \hline\hline
\multirow{2}{0.1cm}{\hspace{-2.1cm}{\textbf{Multi-Layer}}}\hspace{-4mm}\ldelim\{{2}{0.3mm}\hspace{2mm}
\gmn & \second{0.622} & \second{0.71} & 0.73 & 0.662 & \second{0.655} & 0.708 \\[1ex] 
Node pair partner & 0.579 & \first{0.732} & \second{0.74} & 0.677 & 0.641 & \second{0.713}   \\ \hline\hline
\multirow{2}{0.1cm}{\hspace{-2.1cm}{\textbf{Multi-Round}}}\hspace{-4mm}\ldelim\{{2}{0.3mm}\hspace{2mm}
\gmn & \first{0.629} & 0.699 & \first{0.757} & \first{0.697} & 0.653 & \first{0.714} \\[1ex] 
Node pair partner & 0.579 & 0.693 & 0.729 & \second{0.69} & \first{0.665} & 0.705  \\ \hline\hline
 \end{tabular}
}

\bigskip

\maxsizebox{0.9\hsize}{!}{
\small
\begin{tabular}{l|ccc ccc}
\hline
\multicolumn{7}{c}{HITS@20} \\ \hline
& \aids & \mutag & \fm & \fr & \mm & \mr\\ \hline\hline
\multirow{2}{0.1cm}{\hspace{-2.1cm}{\textbf{Multi-Layer}}}\hspace{-4mm}\ldelim\{{2}{0.3mm}\hspace{2mm}
\gmn & \second{0.397} & \second{0.544} & 0.537 & 0.45 & 0.423 & 0.49 \\[1ex] 
Node pair partner & 0.346 & \first{0.567} & \second{0.551} & 0.476 & 0.411 & 0.5  \\ \hline\hline
\multirow{2}{0.1cm}{\hspace{-2.1cm}{\textbf{Multi-Round}}}\hspace{-4mm}\ldelim\{{2}{0.3mm}\hspace{2mm}
\gmn & \first{0.403} & 0.533 & \first{0.562} & \second{0.494} & \second{0.431} & \second{0.502} \\[1ex] 
Node pair partner & 0.344 & 0.528 & 0.54 & \first{0.502} & \first{0.462} & \first{0.506}  \\ \hline\hline
 \end{tabular}
 }

\bigskip

\maxsizebox{0.9\hsize}{!}{
\small
\begin{tabular}{l|ccc ccc}
\hline
\multicolumn{7}{c}{Mean Reciprocal Rank (MRR)} \\ \hline 
& \aids & \mutag & \fm & \fr & \mm & \mr\\ \hline\hline
\multirow{2}{0.1cm}{\hspace{-2.1cm}{\textbf{Multi-Layer}}}\hspace{-4mm}\ldelim\{{2}{0.3mm}\hspace{2mm}
\gmn & \second{0.877} & \first{0.923} & \second{0.949} & \second{0.947} & \second{0.928} & \first{0.922} \\[1ex] 
Node pair partner & 0.827 & 0.897 & \first{0.958} & 0.877 & 0.918 & 0.92  \\ \hline\hline
\multirow{2}{0.1cm}{\hspace{-2.1cm}{\textbf{Multi-Round}}}\hspace{-4mm}\ldelim\{{2}{0.3mm}\hspace{2mm}
\gmn & \first{0.905} & 0.862& \first{0.958} & \first{0.956} & 0.906 & \second{0.921} \\[1ex] 
Node pair partner & 0.811 & \second{0.901} & 0.907 & 0.908 & \first{0.964} & 0.92  \\ \hline\hline
 \end{tabular}
 }

\bigskip

\maxsizebox{0.9\hsize}{!}{
\small
\begin{tabular}{l|ccc ccc}
\hline
\multicolumn{7}{c}{Precision@20} \\ \hline
& \aids & \mutag & \fm & \fr & \mm & \mr\\ \hline\hline
\multirow{2}{0.1cm}{\hspace{-2.1cm}{\textbf{Multi-Layer}}}\hspace{-4mm}\ldelim\{{2}{0.3mm}\hspace{2mm}
\gmn & \second{0.751} & \second{0.82} & 0.852 & 0.809 & 0.783 & 0.832 \\[1ex] 
Node pair partner & 0.7 & \first{0.833} & \second{0.861} & 0.797 & \second{0.792} & \first{0.846}  \\ \hline\hline
\multirow{2}{0.1cm}{\hspace{-2.1cm}{\textbf{Multi-Round}}}\hspace{-4mm}\ldelim\{{2}{0.3mm}\hspace{2mm}
\gmn & \first{0.753} & 0.795 & \first{0.885} & \second{0.829} & \second{0.792} & \second{0.842} \\[1ex] 
Node pair partner & 0.694 & 0.794 & 0.847 & \first{0.835} & \first{0.802} & 0.825 \\ \hline\hline
 \end{tabular}
 }
\caption{\small Effect of node pair partner interaction in \gmn. The tables compare \gmn{} with its \our alternative. The first table reports MAP values, the second table reports HITS@20 values, the third table reports MRR values and the fourth table reports Precision@20. In each table, the first two rows report metrics for multi-layer refinement and the second two rows report metrics for multi-round refinement. Rows colored \fboxg{green} and \fboxy{yellow} indicate the best and second best methods according to the respective metrics.}
\label{tab:cross-our-gmn}
\end{table}




\clearpage
\newpage

\subsection{Variation of \ournode\ and \ouredge\ with different $T$ and $K$}
\label{subsec:inference_time}

In this section, we analyze the accuracy and inference time trade-off of multi-round lazy and multi-layer eager variants of \ournode and \ouredge. In the following tables, we show the MAP and inference time. Additionally, we also analyze the trade-off of \gmn and \isoedge. The $T, K$ parameters for different models are so chosen that they can be compared against each other while fixing the inference time to be roughly similar. For instance, multi-round lazy \ournode with $T=5, K=5$ maps to multi-layer eager \ournode with $K=8$, allowing for a direct comparison of performance without caring much about different compute. Note that in below tables, models are listed in order of increasing inference time (i.e. increasing $K$ or $T$).

In tables ~\ref{tab:t-k-node-T} and ~\ref{tab:t-k-node-K}, we show variations for multi-round lazy \ournode for fixed $T$ and fixed $K$ respectively. We observe that with fixed $T$, increasing $K$ from 5 to 10 doesn't improve the model significantly. For fixed $K$, performance (in terms of MAP) improves notably when increasing $T$ from 3 to 5.

In table ~\ref{tab:k-node}, we show variations for multi-layer eager \ournode for varying $K$. We observe that except for a drop at $K=7$, the performance of the model improves as we increase $K$. In fact, at $K=8$, the performance is surprisingly good, even outperforming the similarly timed $T=5, K=5$ variant of lazy multi-round \ournode on both \aids~and \mutag.

In tables ~\ref{tab:t-k-edge-T} and ~\ref{tab:t-k-edge-K}, we compare variants of multi-round lazy \ouredge with fixed $T$ and fixed $K$ respectively. We observe that when $T$ is fixed and $K$ is increased, the gain is marginal. We observe a significant gain When $K$ is fixed and $T$ is increased from 3 to 4.

In table ~\ref{tab:k-edge}, we study the trade-off for multi-layer eager \ouredge for varying $K$. We observe that with increasing $K$, the performance continues to improve and peaks at $K=8$. Note that even at this $K$, the performance of multi-layer eager \ouredge is worse than a similarly timed variant ($T=5, K=5$) of multi-round \ouredge.

In table ~\ref{tab:k-GMN}, we show variations for \gmn for varying $K$. We observe marginal gains while increasing $K$. From $K=10$ to $K=12$, the performance drops.

In table ~\ref{tab:k-isonet}, we show how performance varies for \isoedge for varying $K$. We observe that the model does not improve with increasing $K$.

\begin{figure}[h]
    \centering
    \vspace{-4mm}

\subfloat[Node, \aids]{\includegraphics[width=.25\linewidth]{FIG/efficiency_p_aids.pdf}}   
\subfloat[Edge, \aids]{\includegraphics[width=.25\linewidth]{FIG/efficiency_s_aids.pdf}}   \hspace{3mm}
\subfloat[Node, \mutag]{\includegraphics[width=.25\linewidth]{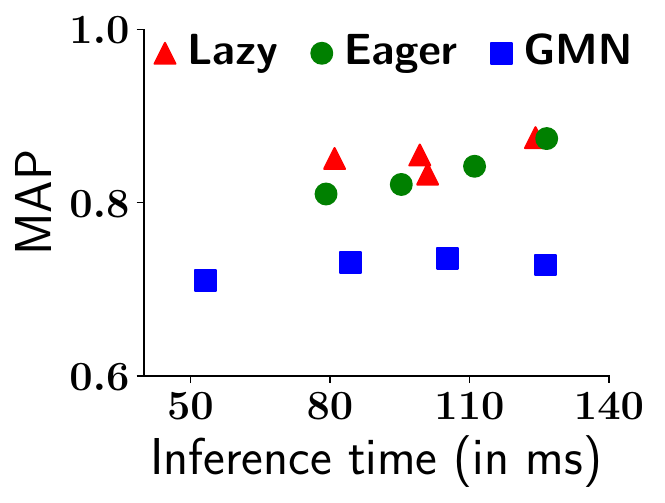}} 
\subfloat[Edge, \mutag]{\includegraphics[width=.25\linewidth]{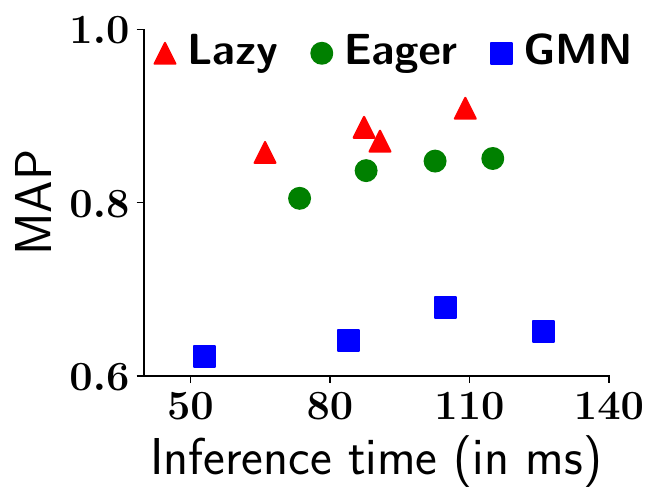}}

\caption{\small Trade off between MAP and inference time (batch size=128).}
\end{figure}

\begin{table}[!!h]
\centering
\maxsizebox{0.9\hsize}{!}{
\small
\begin{tabular}{l|ccc ccc}
\hline
\multicolumn{7}{c}{Mean Average Precision (MAP)} \\ \hline
& \aids & \mutag & \fm & \fr & \mm & \mr \\ \hline\hline
  $T=3,K=5$ & \first{0.825} & 0.851 & 0.888 & \first{0.855} & 0.838 & 0.874\\
  $T=3,K=10$ & 0.774 & \first{0.855} & \first{0.898} & 0.811 & \first{0.855} & \first{0.882} \\ \hline
 \end{tabular}}

\bigskip

\maxsizebox{0.9\hsize}{!}{
\small
\begin{tabular}{l|ccc ccc}
\hline
\multicolumn{7}{c}{Inference time (in ms)} \\ \hline
& \aids & \mutag & \fm & \fr & \mm & \mr \\ \hline\hline
  $T=3,K=5$ & 80.11 & 80.99 & 81.01 & 81.24 & 80.94 & 80.25\\
  $T=3,K=10$ & 99.11 & 99.31 & 99.28 & 99.48 & 99.37 & 99.36 \\ \hline
 \end{tabular}}
\caption{\small MAP and inference time trade-off of variants of multi-round lazy \ournode with fixed $T$. Rows colored \fboxg{green} indicate the best $K$ according to the MAP score.}
\label{tab:t-k-node-T}
\end{table}

\begin{table}[!!h]
\centering
\maxsizebox{0.9\hsize}{!}{
\small
\begin{tabular}{l|ccc ccc}
\hline
\multicolumn{7}{c}{Mean Average Precision (MAP)} \\ \hline

& \aids & \mutag & \fm & \fr & \mm & \mr \\ \hline\hline
  $T=3,K=5$ & \second{0.825} & \second{0.851} & 0.888 & 0.855 & 0.838 & 0.874 \\
  $T=4,K=5$ & 0.799 & 0.833 & \second{0.892} & \second{0.858} & \second{0.867} & \second{0.891} \\
  $T=5,K=5$ & \first{0.845} & \first{0.875} & \first{0.919} & \first{0.883} & \first{0.894} & \first{0.897} \\
  \hline
 \end{tabular}}

\bigskip

 \maxsizebox{0.9\hsize}{!}{
\small
\begin{tabular}{l|ccc ccc}
\hline
\multicolumn{7}{c}{Inference time (in ms)} \\ \hline

& \aids & \mutag & \fm & \fr & \mm & \mr \\ \hline\hline
  $T=3,K=5$ & 80.11 & 80.99 & 81.01 & 81.24 & 80.94 & 80.25\\
  $T=4,K=5$ & 101.33 & 100.99 & 100.95 & 100.46 & 100.59 & 100.87\\
  $T=5,K=5$ & 123.18 & 124.19 & 123.61 & 122.79 & 123.33 & 122.74\\
  \hline
 \end{tabular}}
\caption{\small MAP and inference time trade-off of variants of multi-round lazy \ournode with fixed $K$. Rows colored \fboxg{green} and \fboxy{yellow} indicate the best and second best $T$ according to the MAP score.}
\label{tab:t-k-node-K}
\end{table}

\begin{table}[!!h]
\centering
\begin{minipage}[t]{0.45\linewidth}
\centering
\maxsizebox{\hsize}{!}{
\small
\begin{tabular}{l|cc}
\hline
\multicolumn{3}{c}{Mean Average Precision (MAP)} \\ \hline
& \aids & \mutag\\ \hline\hline
  $K=5$ & 0.756 & 0.81 \\
  $K=6$ & \second{0.813} & 0.821 \\
  $K=7$ & 0.808 & \second{0.842}\\
  $K=8$ & \first{0.883} & \first{0.874}\\
\hline
\end{tabular}}
\end{minipage}
\hspace{10pt} 
\begin{minipage}[t]{0.45\linewidth}
\centering
\maxsizebox{\hsize}{!}{
\small
\begin{tabular}{l|cc}
\hline
\multicolumn{3}{c}{Inference time (in ms)} \\ \hline
& \aids & \mutag\\ \hline\hline
  $K=5$ & 79.02 & 79.15 \\
  $K=6$ & 94.99 & 95.33 \\
  $K=7$ & 110.78 & 111.09 \\
  $K=8$ & 126.48 & 126.6 \\
\hline
\end{tabular}}
\end{minipage}
\caption{\small MAP and inference time trade-off of variants of multi-layer eager \ournode with increasing $K$. Rows colored \fboxg{green} and \fboxy{yellow} indicate the best and second best $K$ according to the MAP score.}
\label{tab:k-node}
\end{table}

\begin{table}[!!h]
\centering
\begin{minipage}[t]{0.45\linewidth}
\centering
\maxsizebox{\hsize}{!}{
\small
\begin{tabular}{l|cc}
\hline
\multicolumn{3}{c}{Mean Average Precision (MAP)} \\ \hline
& \aids & \mutag\\ \hline\hline
  $T=3,K=5$ & 0.847 & 0.858\\
  $T=3,K=10$ & \first{0.865} & \first{0.871}\\ \hline
\end{tabular}}
\end{minipage}
\hspace{10pt} 
\begin{minipage}[t]{0.45\linewidth}
\centering
\maxsizebox{\hsize}{!}{
\small
\begin{tabular}{l|cc}
\hline
\multicolumn{3}{c}{Inference time (in ms)} \\ \hline
  & \aids & \mutag\\ \hline\hline
  $T=3,K=5$ & 64.39 & 66.03\\
  $T=3,K=10$ & 88.59 & 90.76\\ \hline
\end{tabular}}
\end{minipage}
\caption{\small MAP and inference time trade-off of variants of multi-round lazy \ouredge with fixed $T$. Rows colored \fboxg{green} indicate the best $K$ according to the MAP score.}
\label{tab:t-k-edge-T}
\end{table}

\begin{table}[!!h]
\centering
\begin{minipage}[t]{0.45\linewidth}
\centering
\maxsizebox{\hsize}{!}{
\small
\begin{tabular}{l|cc}
\hline
\multicolumn{3}{c}{Mean Average Precision (MAP)} \\ \hline
& \aids & \mutag\\ \hline\hline
  $T=3,K=5$ & 0.847 & 0.858\\
  $T=4,K=5$ & \second{0.881} & \second{0.887}\\
  $T=5,K=5$ & \first{0.886} & \first{0.909}\\
\hline
\end{tabular}}
\end{minipage}
\hspace{10pt} 
\begin{minipage}[t]{0.45\linewidth}
\centering
\maxsizebox{\hsize}{!}{
\small
\begin{tabular}{l|cc}
\hline
\multicolumn{3}{c}{Inference time (in ms)} \\ \hline
& \aids & \mutag\\ \hline\hline
  $T=3,K=5$ & 64.39 & 66.03\\
  $T=4,K=5$ & 85.02 & 87.33\\
  $T=5,K=5$ & 106.24 & 109.1\\
\hline
\end{tabular}}
\end{minipage}
\caption{\small MAP and inference time trade-off of variants of multi-round lazy \ouredge with fixed $K$. Rows colored \fboxg{green} and \fboxy{yellow} indicate the best and second best $T$ according to the MAP score.}
\label{tab:t-k-edge-K}
\end{table}

\clearpage
\newpage

\begin{table}[!!h]
\centering
\begin{minipage}[t]{0.45\linewidth}
\centering
\maxsizebox{\hsize}{!}{
\small
\begin{tabular}{l|cc}
\hline
\multicolumn{3}{c}{Mean Average Precision (MAP)} \\ \hline
& \aids & \mutag\\ \hline\hline
  $K=5$ & 0.795 & 0.805\\
  $K=6$ & 0.828 & 0.837\\
  $K=7$ & \second{0.852} & \second{0.848}\\
  $K=8$ & \first{0.862} & \first{0.851}\\
\hline
\end{tabular}}
\end{minipage}
\hspace{10pt} 
\begin{minipage}[t]{0.45\linewidth}
\centering
\maxsizebox{\hsize}{!}{
\small
\begin{tabular}{l|cc}
\hline
\multicolumn{3}{c}{Inference time (in ms)} \\ \hline
& \aids & \mutag\\ \hline\hline
  $K=5$ & 72.63 & 73.46\\
  $K=6$ & 86.03 & 87.77\\
  $K=7$ & 100.26 & 102.6\\
  $K=8$ & 114.33 & 115.01\\
\hline
\end{tabular}}
\end{minipage}
\caption{\small MAP and inference time trade-off of variants of multi-layer eager \ouredge with increasing $K$. Rows colored \fboxg{green} and \fboxy{yellow} indicate the best and second best $K$ according to the MAP score.}
\label{tab:k-edge}
\end{table}

\begin{table}[!!h]
\centering
\maxsizebox{0.9\hsize}{!}{
\small
\begin{tabular}{l|ccc ccc}
\hline
\multicolumn{7}{c}{Mean Average Precision (MAP)} \\ \hline

& \aids & \mutag & \fm & \fr & \mm & \mr \\ \hline\hline
  $K=5$ & 0.622 & 0.710 & 0.730 & 0.662 & 0.655 & 0.708  \\
  $K=8$ & 0.641 & \second{0.731} & \first{0.745} & \second{0.701} & 0.658 & \second{0.711} \\
  $K=10$ & \first{0.679} & \first{0.736} & 0.741 & \first{0.712} & \first{0.691} & \first{0.74} \\
  $K=12$ & \second{0.651} & 0.728 & \second{0.743} & 0.697 & \second{0.687} & 0.699\\
  \hline
 \end{tabular}}

\bigskip

 \maxsizebox{0.9\hsize}{!}{
\small
\begin{tabular}{l|ccc ccc}
\hline
\multicolumn{7}{c}{Inference time (in ms)} \\ \hline

& \aids & \mutag & \fm & \fr & \mm & \mr \\ \hline\hline
$K=5$ & 52.94 & 53.16 & 53.23 & 53.12 & 53.32 & 53.34\\
  $K=8$ & 83.97 & 84.47 & 84.64 & 84.38 & 85.41 & 84.51\\
  $K=10$ & 104.87 & 105.21 & 105.72 & 105.33 & 105.66 & 105.73\\
  $K=12$ & 125.99 & 126.33 & 126.53 & 126.39 & 126.79 & 126.59\\
  \hline
 \end{tabular}}
\caption{\small MAP and inference time trade-off of variants of \gmn with increasing $K$. Rows colored \fboxg{green} and \fboxy{yellow} indicate the best and second best $K$ according to the MAP score.}
\label{tab:k-GMN}
\end{table}

\begin{table}[!!h]
\centering
\maxsizebox{0.9\hsize}{!}{
\small
\begin{tabular}{l|c|c}
\hline
& \aids & Inference time (in ms)\\ \hline\hline
  $K=5$ & 0.69 & 19.77\\
  $K=6$ & \first{0.717} & 20.83 \\
  $K=7$ & 0.697 &  21.96\\
  $K=8$ & \second{0.709} & 23.02 \\
  \hline
 \end{tabular}}
\caption{\small MAP and inference time trade-off of variants of \isoedge with increasing $K$. Rows colored \fboxg{green} and \fboxy{yellow} indicate the best and second best $T$ according to the MAP score.}
\label{tab:k-isonet}
\end{table}


\subsection{Contribution of refining alignment matrix in inference time}
In GMN, computing the embeddings of nodes after the message passing step at each layer dominates the inference time. However, in the case of \our models, we observe the refinement of the alignment matrix at each layer or round to also be time-intensive. 
In table ~\ref{tab:embedding-matrix-time}, we show the contribution of embedding computation and matrix updates to the total inference time. The updates to $P$ constitute the largest share of inference time for multi-layer variants. This can be attributed to the refinement of $P$ after every message passing step, equaling the frequency of embedding computation. In the case of multi-round variants, both embedding computation and updates to $P$ contribute almost equally since $P$ is refined only at the end of each round, after several layers of message passing alongwith embedding computation.

\begin{table}[!!h]
\centering
\maxsizebox{0.9\hsize}{!}{
\small
\begin{tabular}{l|c|c}
\hline
Models & Embedding Computation & Matrix Updates\\ \hline\hline
  multi-layer \ournode & 13.7 & 68.3\\
  multi-layer \ouredge & 19.7 & 76.3 \\
  multi-round \ournode & 34.1 &  47.8\\
  multi-round \ouredge & 54.9 & 39.9 \\
  \hline
 \end{tabular}}
\caption{\small Inference time contribution of embedding computation and matrix updates by multi-layer and multi-round \ournode and \ouredge models.}
\label{tab:embedding-matrix-time}
\end{table}


\subsection{Transfer ability of learned models}
In this section, we evaluate the transfer ability of each trained model across datasets. In table~\ref{tab:transfer-mutag}, we report the Mean Average Precision (MAP) scores for models trained using the AIDS and Mutag datasets respectively evaluated on all six datasets. We observe that despite a zero-shot transfer from one of the datasets to all others, variants of \our show the best accuracy.

\begin{table}[h!]
\centering
\maxsizebox{0.7\hsize}{!}{  \tabcolsep 7pt 
\begin{tabular}{l|c c c c c c }
\toprule
\multicolumn{7}{c}{Test across other datasets when trained on \textbf{\aids}} \\ \hline
& \aids & \mutag & \fm & \fr& \mm & \mr  \\
\midrule \midrule 
\graphsim~\cite{graphsim}  & 0.356 & 0.225 & 0.192 & 0.198 & 0.210 & 0.215 \\[1ex] 
\gotsim~\cite{gotsim} & 0.324 & 0.275 & 0.370 & 0.339 & 0.314 & 0.361\\[1ex] 
\simgnn~\cite{simgnn}  & 0.341 & 0.264 & 0.374 & 0.344 & 0.331 & 0.383\\[1ex] 
\egsc~\cite{egsc}  & 505 & 0.255 & 0.473 & 0.451 & 0.447 & 0.499\\[1ex] 
\hmn~\cite{zhang2021h2mn}  & 0.267 & 0.272 & 0.319 & 0.281 & 0.262 & 0.297\\ [1ex] \hline 
\neuromatch~\cite{neuromatch}  & 0.489 & 0.287 & 0.442 & 0.403 & 0.386 & 0.431\\[1ex] 
\greed~\cite{greed}  & 0.472 & 0.307 & 0.477 & 0.452 & 0.436 & 0.490\\[1ex] 
\gen~\cite{gmn}  & 0.557 & 0.291 & 0.445 & 0.427 & 0.437 & 0.496\\[1ex] 
\gmn~\cite{gmn}  & 0.622 & 0.342 & 0.569 & 0.544 & 0.532 & 0.588\\[1ex] 
\isonode~\cite{RoyVCD2022IsoNet}  & 0.659 & 0.459 & 0.612 & 0.562 & 0.588 & 0.640\\[1ex] 
\isoedge~\cite{RoyVCD2022IsoNet}  & 0.690 & 0.468 & 0.620 & 0.568 & 0.624 & 0.627 \\[1ex] 
\hline
 \ournodelayershort & 0.756 & 0.685 & 0.825 & 0.767 & 0.781 & 0.794  \\[1ex] 
\ouredgelayershort & 0.795 & 0.683 & 0.800 & 0.751 & 0.792 & 0.785  \\[1ex]  \hline
\ournoderoundshort &\second{0.825} & \second{0.702} & \second{0.828} & \second{0.777} & \second{0.800} & \second{0.825}\\[1ex] 
\ouredgeroundshort & \first{0.847} & \first{0.741} & \first{0.846} & \first{0.799} & \first{0.833} & \first{0.836}\\  \hline\hline 

\multicolumn{7}{c}{Test across other datasets when trained on \textbf{\mutag}} \\ \hline
& \aids & \mutag & \fm & \fr& \mm & \mr  \\
\midrule \midrule 
\graphsim~\cite{graphsim} & 0.188 & 0.472 & 0.190 & 0.193 & 0.205 & 0.198\\[1ex] 
\gotsim~\cite{gotsim} & 0.194 & 0.272 & 0.185 & 0.192 & 0.202 & 0.182\\[1ex] 
\simgnn~\cite{simgnn} & 0.206 & 0.283 & 0.203 & 0.209 & 0.220 & 0.195\\[1ex] 
\egsc~\cite{egsc} &0.296 & 0.476 & 0.391 & 0.333 & 0.309 & 0.355 \\[1ex] 
\hmn~\cite{zhang2021h2mn} & 0.209 & 0.276 & 0.204 & 0.207 & 0.223 & 0.197 \\ [1ex] \hline 
\neuromatch~\cite{neuromatch} & 0.275 & 0.576 & 0.368 & 0.304 & 0.304 & 0.325\\[1ex] 
\greed~\cite{greed} & 0.328 & 0.567 & 0.388 & 0.335 & 0.356 & 0.370\\[1ex] 
\gen~\cite{gmn} & 0.278 & 0.605 & 0.359 & 0.308 & 0.312 & 0.330\\[1ex] 
\gmn~\cite{gmn} & 0.299 & 0.710 & 0.434 & 0.361 & 0.389 & 0.394\\[1ex] 
\isonode~\cite{RoyVCD2022IsoNet} & 0.458 & 0.697 & 0.503 & 0.456 & 0.446 & 0.486\\[1ex] 
\isoedge~\cite{RoyVCD2022IsoNet} & 0.472 & 0.706 & 0.499 & 0.438 & 0.467 & 0.489\\ [1ex] 
\hline
\ournodelayershort &  0.601 & 0.810 & \first{0.695} & \second{0.611} & 0.628 & 0.614\\[1ex] 
\ouredgelayershort & 0.527 & 0.805 & 0.558 & 0.507 & 0.560 & 0.563 \\ [1ex] \hline
\ournoderoundshort & \first{0.645} & \second{0.851} & \second{0.679} & \first{0.626} & \first{0.652} & \first{0.655}\\[1ex] 
\ouredgeroundshort &  \second{0.625} & \first{0.858} & 0.639 & 0.598 & \second{0.634} & \second{0.650} \\[1ex]
 \hline \hline  
 \end{tabular}
 }
 \vspace{1mm}
\caption{Test MAP of all graph retrieval methods on different datasets, when they were trained on \textbf{\aids} (top half) and \textbf{\mutag} (bottom half) dataset. The numbers with \fboxg{green} and \fboxy{yellow} indicate the best, second best method respectively.}
\label{tab:transfer-mutag}
\end{table}

\end{document}